\documentclass{article}

\usepackage{natbib}
\setcitestyle{authoryear}

\usepackage[final]{neurips_2025}

\usepackage[utf8]{inputenc} 
\usepackage[T1]{fontenc}    
\usepackage{hyperref}       
\usepackage{url}            
\usepackage{booktabs}       
\usepackage{amsfonts}       
\usepackage{nicefrac}       
\usepackage{microtype}      
\usepackage{xcolor}         

\definecolor{bluecite}{HTML}{0875b7}
\hypersetup{
  colorlinks=true,
  citecolor=bluecite,
  linkcolor=magenta
}

\usepackage{graphicx}
\usepackage{wrapfig}
\usepackage{caption}
\usepackage{subcaption}
\usepackage{amsmath,amssymb}
\usepackage{array,multirow}
\usepackage{xspace}
\usepackage{tabularx}
\usepackage{float}
\usepackage{siunitx}
\usepackage[capitalise]{cleveref}
\usepackage{enumitem}

\usepackage{algorithm}
\usepackage[noend]{algorithmic}

\newcommand{\compass}{{\sc compass}\xspace}
\newcommand{\sable}{{\sc sable}\xspace}
\newcommand{\ippo}{{\sc ippo}\xspace}
\newcommand{\mappo}{{\sc mappo}\xspace}
\newcommand{\sgbs}{{\sc sgbs}\xspace}
\newcommand{\cmaes}{{\sc cma-es}\xspace}

 \newcommand{\del}[1]{}

\title{Breaking the Performance Ceiling in Reinforcement Learning requires Inference Strategies}

\author{
Felix Chalumeau\thanks{Equal contribution. Corresponding author: \texttt{f.chalumeau@instadeep.com}}~~$^1$~~~\textbf{Daniel Rajaonarivonivelomanantsoa}$^*$$^{1,2}$~~~\textbf{Ruan de Kock}$^*$$^1$\\ \textbf{Claude Formanek}$^1$~~~\textbf{Sasha Abramowitz}$^1$~~~\textbf{Omayma Mahjoub}$^1$~~~\textbf{Wiem Khlifi}$^1$\\ \textbf{Simon Du Toit}$^1$~~~~\textbf{Louay Ben Nessir}$^1$~~~~\textbf{Refiloe Shabe}$^1$~~~~\textbf{Noah De Nicola}$^1$\\ \textbf{Arnol Fokam}$^1$~~~~\textbf{Siddarth Singh}$^1$~~~~~\textbf{Ulrich Mbou Sob}$^1$~~~~~\textbf{Arnu Pretorius}$^{1,2}$ \\
\\
$^1$InstaDeep \\
$^2$Stellenbosch University
}

\begin{document}

\maketitle

\begin{abstract}

Reinforcement learning (RL) systems have countless applications, from energy-grid management to protein design. However, such real-world scenarios are often extremely difficult, combinatorial in nature, and require complex coordination between multiple agents. This level of complexity can cause even state-of-the-art RL systems, trained until convergence, to hit a performance ceiling which they are unable to break out of with zero-shot inference. Meanwhile, many digital or simulation-based applications allow for an inference phase that utilises a specific time and compute budget to explore multiple attempts before outputting a final solution. In this work, we show that such an inference phase employed at execution time, and the choice of a corresponding inference strategy, are key to breaking the performance ceiling observed in complex multi-agent RL problems. Our main result is striking: \textbf{we can obtain up to a 126\% and, on average, a 45\% improvement over the previous state-of-the-art across 17 tasks, using only a couple seconds of extra wall-clock time during execution}. We also demonstrate promising compute scaling properties, supported by over 60k experiments, making it the largest study on inference strategies for complex RL to date. \textbf{Our experimental data and code are available at \url{https://sites.google.com/view/inference-strategies-rl}}.

\end{abstract}

\begin{figure}[ht]
    \centering
    \includegraphics[width=0.82\textwidth]{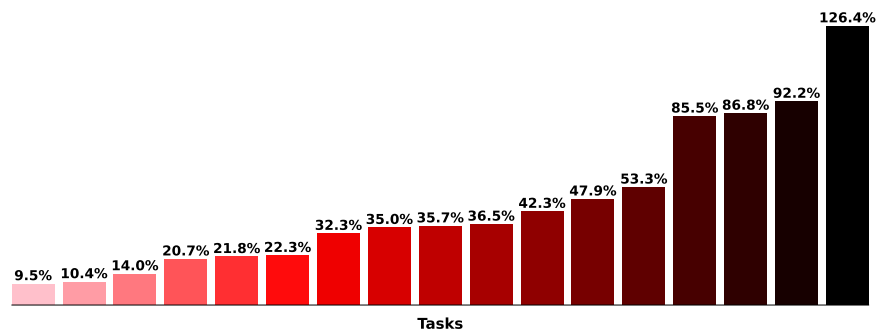}
    \caption{\textbf{Improvement from using inference-time search over zero-shot state-of-the-art}. Across 17 complex reinforcement learning tasks, we obtain consistent and significant performance gains using only a 30 second search budget during execution.}
    \label{fig:main-figure}
\end{figure}

\section{Introduction}

Learning to solve sequential decision-making tasks is a central challenge in artificial intelligence (AI), with far-reaching applications ranging from energy-grid optimisation~\citep{AHMAD2021125834} and autonomous logistics~\citep{laterre2018rankedrewardenablingselfplay} to molecular discovery~\citep{olivecrona2017molecular} and drug design~\citep{popova2018deep}. Complex sequential real-world problems that cannot be solved by traditional optimisation techniques are inherently complex and require navigating high-dimensional solution spaces. Reinforcement Learning (RL) presents a promising avenue to improve our capacity to find efficient solutions to these problems, but despite stunning progress over the past decade, such as human-level performance in Atari games~\citep{mnih2015dqn}, defeating the world champion in the game of Go~\citep{silver2016mastering} or aligning AI systems with human preferences~\citep{stiennon2020summarizehf}, current approaches are facing challenges that prevent their common deployment in most real-world systems~\citep{dulac2020empirical}.

A major source of this difficulty lies in the combinatorial nature of many decision-making tasks. As the problem size increases, the space of possible solutions grows exponentially~\citep{karp1975computational}. In multi-agent systems, the challenge compounds: agents must coordinate in environments where only partial information is available, the joint action space is combinatorial, and optimal behaviour depends on precise interaction with other agents~\citep{bernstein2000complexity,canese2021marl}. These properties make it fundamentally difficult to rely on the zero-shot performance of a trained policy, even if that policy was optimised to convergence on a representative training distribution. This causes the gap between zero-shot performance and optimality to grow substantially with increasing complexity (see~\cref{fig:main-figure}).

However, numerous practical applications are not restricted to producing a single zero-shot solution. Instead, inference is often permitted to take place over a few seconds, minutes or hours, with a given computational resource. Furthermore, models and simulators are often accessible and very efficient (e.g., energy grid management, train scheduling, package delivery, routing, printed circuit board design) and provide either an exact score or a very accurate approximation. In other applications where the gap to reality may be larger (e.g., protein design, robotics), improving the solution under the simulated score can still arguably provide significant improvement towards the real objective~\citep{esm3,dona2024qualitydiversity,hundt2019goodrobot,rao2020rlcyclegan}.

This opens up an opportunity: rather than relying on a single attempt of the trained policy, the time budget and compute capacity can be leveraged to actively search for better solutions using multiple attempts, following an \textit{inference-time} strategy. For instance, progressively building a tree of possible solutions, or adapting the policy using outcomes of past attempts. Even straightforward strategies can provide significant performance improvement with low time cost. For instance, generating a large batch of diversified solutions in parallel, using stochastic sampling, rather than a single greedy solution: given a modern GPU, this enables to produce hundreds of solutions, for the same wall-clock time, enabling massive exploration at no time cost.

These strategies are rarely emphasized in existing benchmarks~\citep{papoudakis2021benchmarking,mahjoub2025sable}, and many practitioners invest months of research trying to improve the zero-shot performance of their models on scenarios where only marginal improvement may still be achieved. Whereas they could unlock performance gains from inference-time search, at negligible wall-clock time cost with moderate compute capacity (\cref{fig:main-figure}).

Research in RL for Combinatorial Optimisation (CO) has produced efficient inference strategies~\citep{Bello16,hottung2022efficient,choo2022simulationguided,chalumeau2023combinatorial}, often referred to as active search, or online adaptation methods. However, their empirical study is still limited to a few problems, a narrow range of budget settings~\citep{chalumeau2024memento}, and barely no insight on their scaling properties. In the multi-agent case, there is no study on inference strategies for collaborative teams: most work on team adaptation focus on Ad Hoc Teamwork~\citep{mirsky2022adhoc,wang2024nagent,ruhdorfer2025overcooked}, which is adjacent to our objective. Interestingly, most recent studies about the impact of inference strategies come from the Large Language Model (LLM) literature~\citep{snell2025scaling,muennighoff2025s1simpletesttimescaling,wu2025inferencescalinglawsempirical}, where the adequate combination of efficient models with inference strategies is currently state-of-the-art (SOTA).

In this work, we formalise and investigate the role of inference strategies in complex decision-making tasks. To capture the full complexity of tasks described above, we formulate our problem setting as a decentralised partially observable Markov decision process (Dec-POMDP) \citep{kaelbling1998planning}. This is instead of the typical single-agent MDP used in many RL studies. We make this choice for several reasons: (1) it more accurately maps onto many complex real-world problems of interest, (2) Dec-POMDPs subsume MDPs by being strictly more complex~\citep{bernstein2000complexity}, and (3) because of this, we expect our findings to translate to all simpler problem formulations. Within this setting, we provide a unifying view of popular inference-strategy paradigms, including policy sampling, tree search~\citep{choo2022simulationguided}, online fine-tuning~\citep{Bello16,hottung2022efficient}, and diversity-based search~\citep{chalumeau2023combinatorial}. Strikingly, we show that across a wide range of specifically selected difficult RL problems, inference strategies boost performance on average by \num{45}\%, over zero-shot SOTA. Furthermore, in the best of cases, this boost can be as large as \num{126}\%. All of this, using only a couple of seconds of additional execution time. 

Our results call for a shift in how RL systems are evaluated and deployed: inference strategies are not a minor post-processing step, but a key performance driver in realistic conditions. This work sets the foundation for a more nuanced view of inference in sequential decision-making and provides the tools to build systems that can scale with compute. All our code and experimental data can be accessed at: \url{https://sites.google.com/view/inference-strategies-rl}.


\section{Related work}
\label{related_work}

\paragraph{Inference Strategies from RL for CO} Beyond naive stochastic sampling, several paradigms have been explored to generate the best possible solution using a trained policy checkpoint during inference. Online fine-tuning~\citep{Bello16} retrains all policy parameters with RL using past attempts. \citet{hottung2022efficient} re-trains only a subset of the policy's parameters to reduce memory and compute overheads, enabling more attempts for a given inference budget, and adds an imitation learning term to the RL term, to force exploration close to the best solution found so far. \citet{macfarlane2024spo} also investigates inference-time policy improvement via sequential updates. \citet{choo2022simulationguided} uses tree search, with simulation guided node estimates under budget constraints, which outperforms Beam Search and Monte Carlo Tree Search~\citep{coulom2006mcts}. Diversity-based methods: inspired by previous diversity-seeking approaches, like unsupervised skill discovery \citep{eysenbach2018diversity,sharma2019dynamics,kumar2020one} and quality-diversity \citep{chalumeau2023neuroevolution,cully2017quality}. \citet{poppy} introduces an RL objective that trains a population of diverse and specialized policies, efficient for few-shot performance. \citet{chalumeau2023combinatorial} uses this objective and encodes the diversity in a continuous latent space that can be searched at inference-time, introducing the SOTA method \compass; meanwhile~\citet{hottung2024polynet} uses a similar approach but with a discrete encoding space.

These works have introduced most of the inference strategies we consider in this paper, but they fall short on three important aspects that we aim to improve: (i) they evaluate inference strategies on benchmarks where over \num{95}\% zero-shot optimality is already achieved, leaving little room for meaningful gains, (ii) these benchmarks rely on domain-specific tricks such as starting points or instance augmentations; and (iii) methods are compared under a unique budget setting, overlooking the fact that relative performance depends on the available compute and time budget. In addition, their ability to scale with compute remains unexplored, despite being a critical property.

\paragraph{Policy adaptation in Meta-RL and Offline-to-Online RL} Meta-RL and offline-to-online RL both design mechanisms for policy adaptation, thereby sharing close conceptual links with several popular inference strategies. \compass~\citep{chalumeau2023combinatorial} and VariBAD~\citep{zintgraf2021varibad} both condition a policy on a latent space and search it to adapt. MEMENTO~\citep{chalumeau2024memento} and RL$2$~\citep{duan2016rl2fastreinforcementlearning} similarly rely on memory and a learned rule to search the policy space. DIMES~\citep{qiu2022dimes} explicitly performs meta-RL following MAML’s methodology~\citep{finn2017maml}. While we are not aware of inference strategies directly inspired by the offline-to-online RL literature, \citet{nakamoto2023calql} and \citet{mark2025policyagnostic} introduce mechanisms which could be beneficial at inference time.

It is worth highlighting fundamental distinctions between these fields. Inference strategies focus on (i) the solutions found rather than the learned policy, (ii) maximum rather than average performance, and (iii) single instances at test time, whereas most meta-RL and offline-to-online RL methods adapt from a distribution to another distribution to improve generalisation.

\paragraph{Search and adaptation in Multi-Agent RL} There is only limited work on inference strategies for MARL. Most work about search and adaptation within MARL focus on the challenge of Ad Hoc Teamwork~\citep{yourdshahi2018largescaleadhoc,hu2020otherplay,mirsky2022adhoc,wang2024nagent,ruhdorfer2025overcooked,hammond2025symmetrybreaking}, often in the form of zero-shot coordination, where agents must generalize to new partners at execution time. While these lines of work share some methodological similarities, for instance using diversity-seeking training~\citep{long2024roleplaylearningadaptive,lupu2021trajectory} or adapting through tree search~\citep{yourdshahi2018largescaleadhoc}, they pursue fundamentally different goals and remain orthogonal. In our work, our focus is primarily on solving difficult and complex industrial optimisation tasks.

\paragraph{Inference-time compute for LLMs} Recent advances in LLMs are closely intertwined with the use of inference strategies~\citep{snell2025scaling,wei2022chainofthoughts,wang2024hypothesis}, and a growing effort has gone into studying their scaling properties~\citep{muennighoff2025s1simpletesttimescaling,wu2025inferencescalinglawsempirical}. However, the typical inference-time setting is usually different from ours. LLMs have very costly forward passes, and cannot access the exact score of their answers, but can approximate them using a reward model~\citep{ouyang2022rlhf}. Most popular strategies for LLMs are designed for few shots, namely sampling and ensembling (e.g., majority voting).

Overall, numerous efficient inference strategies have been proposed in the literature, yet their efficiency under various settings remains unexamined. Multi-agent RL, despite its inherent complexity, rarely considers inference-time search beyond Ad Hoc Teamwork. Moreover, the broader field of decision-making has not systematically studied how inference strategies scale with compute. Our work aims at filling these gaps by extending the evaluation of inference strategies in RL, demonstrating major performance gains over a wide range of budget settings with impressive scaling properties.
\begin{figure}[t]
    \centering
    \includegraphics[width=0.98\textwidth]{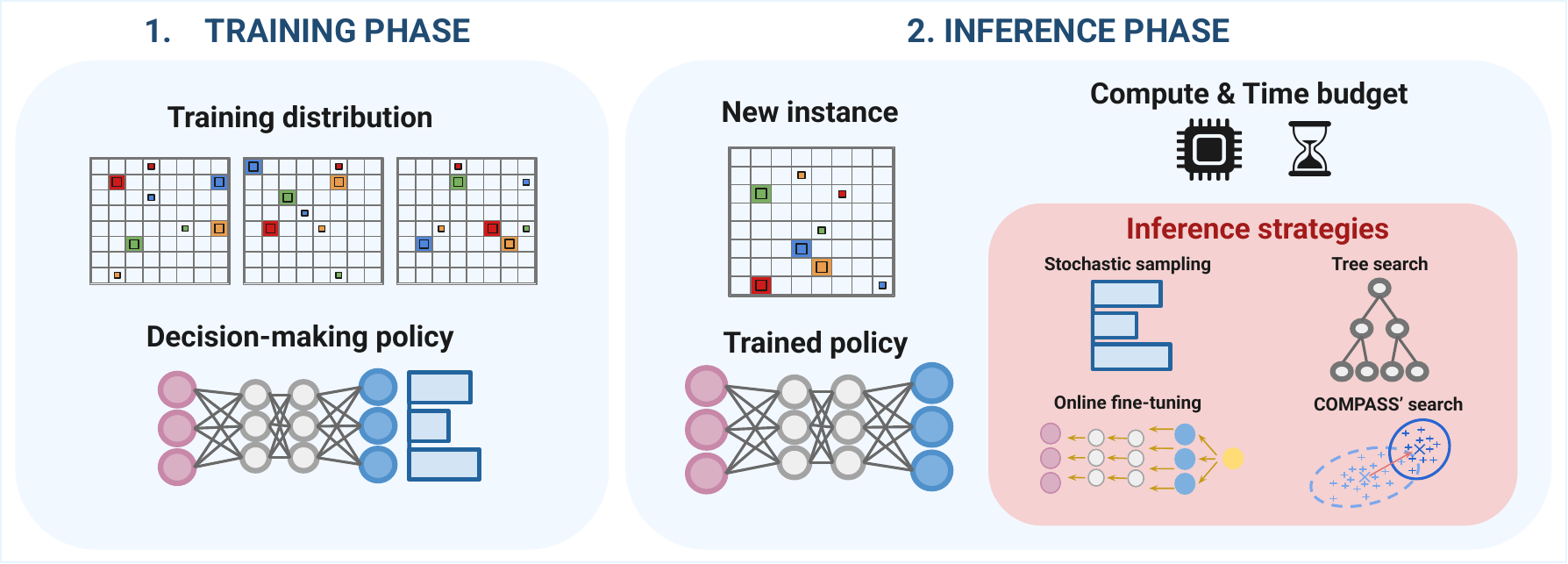}
    \caption{\textbf{Numerous applications of RL involve two distinct phases:} (1) a training phase, typically unconstrained in time and compute, during which a policy is optimized over a representative distribution of problem instances; and (2) an inference phase, where a limited time and compute budget are allocated to solving a new instance. The inference phase is often overlooked, despite its crucial role in complex tasks where partial observability and the combinatorial growth of observation and action spaces make good solutions unattainable through zero-shot execution alone.
}
    \label{fig:method}
\end{figure}

\section{Finding the best solution for a given time and compute budget}
\label{method}

\subsection{Preliminaries}

We focus on RL approaches, and use a neural network (policy) that can construct a solution by taking a sequence of actions. This policy is optimised during a training phase and then used during an inference phase, along with an \textit{inference strategy}, to construct the best possible solution to a new problem instance under a given time and compute budget. These two phases, illustrated in~\cref{fig:method}, have different assumptions, objectives and constraints, detailed in the following paragraphs.

\paragraph{Problem instances} 
We assume that each problem instance can be formulated as a Dec-POMDP \citep{kaelbling1998planning}, defined by the tuple
  $\mathcal{M}
  = \bigl(
      N,\,
      \mathcal{S},\,
      \boldsymbol{\mathcal{O}},\,
      \boldsymbol{\Omega},\,
      \boldsymbol{\mathcal{A}},\,
      R,\,
      P,\,
      \gamma,\,
      H
    \bigr)$.
Here \(N\) is the number of agents, \(\mathcal{S}\) the environment state space, \(\boldsymbol{\mathcal{O}} = \prod_{i=1}^{N}\mathcal{O}^{i}\) the joint agent observation space, \(\boldsymbol{\Omega} : \mathcal{S} \mapsto \boldsymbol{\mathcal{O}}\) the observation function, \(\boldsymbol{\mathcal{A}} = \prod_{i=1}^{N}\mathcal{A}^{i}\) the joint action space, \(R : \mathcal{S} \times \boldsymbol{\mathcal{A}} \mapsto \mathbb{R}\) the shared reward function, \(P : \mathcal{S} \times \boldsymbol{\mathcal{A}} \mapsto \Delta \mathcal{S} \) the environment transition function, the scalar \(\gamma \in [0,1]\) is the discount factor and \(H\) the finite episode horizon. At each timestep, the environment occupies a state \(s_t\) which is mapped to a joint partial observation \( \boldsymbol{o}_t\) via $\boldsymbol{\Omega}$. The joint action is then sampled following the joint policy, \(\boldsymbol{a}_t \sim \boldsymbol{\pi}(\,\cdot \mid \boldsymbol{o_t})\) and is executed; after which the environment transitions to the next state \(s_{t+1}\), following \(P(\,\cdot \mid s_t,\boldsymbol{a}_t)\), and the team receives a reward \(r_t = R(s_t,\boldsymbol{{a}_t})\).

\paragraph{Training Phase} We assume a distribution of problem instances \(\mathcal{D}\), that can be sampled from during training. The joint policy $\boldsymbol{\pi}_{\boldsymbol{\theta}}$, parameterised by $\boldsymbol{\theta}$, is used to construct a solution sequentially by taking joint actions conditioned on the joint observation at each timestep. We use RL to train this policy to maximise the expected return obtained when building solutions to instances drawn from the distribution \(\mathcal{D}\) over a horizon $H$:
\(
J(\boldsymbol{\pi}_{\boldsymbol{\theta}}) = \mathbb{E}_{\mathcal{D}} \left[ \sum_{t=0}^{H} \gamma^t R(s_t, \boldsymbol{a_t}) \right]
\). 

This training objective corresponds to a single attempt (zero-shot). Ideally, this objective should anticipate the multiple attempts allowed at inference, but this is hard to scale. Recent works incorporate such few-shot objectives~\citep{poppy,chalumeau2023combinatorial,chalumeau2024memento}, but none can yet scale beyond \num{200} attempts.

The training phase is usually loosely constrained in terms of time and compute capacity, as typically industrial stakeholders are willing to invest days, weeks or even months of training to obtain a high-performing policy that can generate accurate solutions when deployed in production. Hence, in our experiments, we train all policies until convergence.

\paragraph{Inference Phase} At inference time, a new problem instance \(\rho\) is drawn from a distribution \(\mathcal{D'}\) (possibly different than \(\mathcal{D}\)). Here, there are typically hard constraints to outputting a final solution: a fixed time limit \(\text{T}_{\text{max}}\) constrains wall-clock execution, and a compute capacity \(\text{B}_{\text{max}}\) constrains the number of operations that can be done in parallel. The trained policy \(\boldsymbol{\pi}_{\boldsymbol{\theta}}\) can be used within these constraints to generate solutions to the problem, and the best solution is ultimately used. The reward function \(R\) can still be used to score attempted solutions and inform subsequent attempts. \textit{Inference strategies} can be defined as a function  
\(
\mathcal{I}: (\rho, \boldsymbol{\pi}_{\boldsymbol{\theta}}, \text{B}_{\text{max}}, \text{T}_{\text{max}}) \mapsto (\boldsymbol{a}_1^*, ..., \boldsymbol{a}_H^*)
\)
that uses the base policy \(\boldsymbol{\pi}_{\boldsymbol{\theta}}\) and any additional inference-time search, adaptation, storage, or optimisation methods under the budget \(\text{T}_{\text{max}}\) and \(\text{B}_{\text{max}}\) to produce the best possible solution to the problem instance \(\rho\), defined by the sequence of actions \((\boldsymbol{a}_i^*)_{1 \leq i \leq H}\). The objective can hence be written as:  
\[
I(\mathcal{I}) = \sum_{t=0}^{H} R(s_t, \boldsymbol{a}_t^*) ~~ \text{s.t.} ~~ \text{C}(\mathcal{I}) \leq \text{B}_{\text{max}},~ \text{T}(\mathcal{I}) \leq \text{T}_{\text{max}}
\]
where \(\text{C}(\mathcal{I})\) and \(\text{T}(\mathcal{I})\) represent the compute and time cost of the inference strategy. This formulation highlights that, unlike traditional RL, where zero-shot performance is the primary measure, we focus on strategies which enable further improvement under given constraints. We provide three real-world examples in~\cref{app:real-world} that correspond to this problem setting, illustrating practical scenarios where inference-time search is both natural and feasible.

Inference strategies differ in how they explore the solution space and how they incorporate the outcomes of previous attempts to influence future sampling. Their effectiveness depends on several contextual factors, including the parameter count of the pre-trained policy, the problem’s underlying structure, the episode horizon, the nature of the reward function, but most critically, on the available time and compute budget.

\subsection{One budget, many possibilities: inference-time search and adaptation}
\label{sec:main-methods}

In this section, we detail four types of inference strategies and how we adapt them to work in the multi-agent setting. We implement and release all of these methods in JAX \citep{jax2018github}.

\paragraph{Stochastic policy sampling} The first natural lever to improve solution quality is to re-sample from a stochastic policy. In other words, beyond the creation of a unique greedy solution (i.e., using \(a = \arg\max_{a'} \pi_{\theta}(a'|o)\) over a trajectory of observations), one can sample stochastically (as in \(a \sim \pi_{\theta}(\cdot|o)\)) in order to create diverse solutions. \textbf{Multi-agent policy sampling} generalises easily to the multi-agent case by sampling from the joint action distribution, $\boldsymbol{a} \sim \boldsymbol{\pi}_{\boldsymbol{\theta}}(\cdot|\boldsymbol{o})$.

\paragraph{Tree search} These methods store information about partial solutions using past attempts to preferentially search promising regions of the solution space without updating the pre-trained policy. Simulation guided beam search (\sgbs)~\citep{choo2022simulationguided} provides the best time to performance balance in the literature, outperforming Monte Carlo Tree Search~\citep{coulom2006mcts}. Like most tree searches, \sgbs has three steps: expansion, simulation, pruning. Expansion uses the policy to decide on the most promising next actions (i.e., \(a=\text{top-}K(\pi_{\theta}(\cdot|o))\) ) from the current node (partial solution). A simulated rollout of an episode is produced greedily using \(\pi_{\theta}\) and the return is collected for each node. Pruning keeps only the best nodes found so far based on the return. Solely the expansion step needs to be adapted for \textbf{Multi-agent SGBS}. This is trivial when the explicit joint actions are accessible~\citep{dewitt2020independentlearningneedstarcraft,yu2022mappo}, since we can still select the top ones (i.e., \(\boldsymbol{a}=\text{top-}K(\boldsymbol{\pi}_{\boldsymbol{\theta}}(\cdot|\boldsymbol{o}))\)). For methods using auto-regressive action selection~\citep{mahjoub2025sable}, having access to the top joint actions is intractable, hence we sample K times stochastically from the same node (i.e., \(\boldsymbol{a}[1], ..., \boldsymbol{a}[K] \sim \boldsymbol{\pi}_{\boldsymbol{\theta}}(\cdot|\boldsymbol{o})\)).

\paragraph{Online fine-tuning} These methods keep updating policy parameters at inference time. Given a base policy \(\pi_{\theta}\), online fine-tuning optimises \(\theta\) using inference-time rollouts and policy gradient updates:  
\(
\theta' = \theta + \alpha \nabla_{\theta} J(\pi_{\theta})
\),
where \(J(\pi_{\theta})\) represents an adaptation objective. In line with~\citep{Bello16}, we keep maximising expected returns~\citep{Bello16} over past attempts on the fixed instance (instead of over a training distribution). \textbf{Multi-agent online fine-tuning} re-trains $\boldsymbol{\pi}_{\boldsymbol{\theta}}$ on the new instance using the MARL algorithm that was used during pre-training~\citep{Bello16,mahjoub2025sable,dewitt2020independentlearningneedstarcraft,yu2022mappo}.

\paragraph{Diversity-based approaches} These methods pre-train a collection of diverse specialised policies which can be used to search for the most appropriate solution at inference-time. \compass~\citep{chalumeau2023combinatorial} encodes specialised policies in a continuous latent space \(\mathcal{L}\) by augmenting a pre-trained policy to condition on both the observation and a latent vector sampled from $\mathcal{L}$ (i.e., \(a \sim \pi_\theta(\,\cdot \mid o, ~z), ~~ z \sim \mathcal{L}\)): effectively creating a continuous collection of policies. \compass achieves SOTA in single-agent RL for CO. To avoid having the latent space of \textbf{Multi-agent COMPASS} growing exponentially with the number of agents, we keep one latent space \(\mathcal{L}\) for all agents (i.e., \( \boldsymbol{a} \sim \boldsymbol{\pi}_{\boldsymbol{\theta}}(\,\cdot \mid \boldsymbol{o}, ~z), ~~ z \sim \mathcal{L}
\)). This allows for tractable training, and for efficient inference search with the covariance matrix adaptation evolution strategy \cmaes~\citep{hansen2001cmaes}. Aside from being multi-agent, we keep the training and inference phases close to the original method described in~\citet{chalumeau2023combinatorial}, and provide further details and explanations in~\cref{app:compass-details}.

Unlike other inference strategies, \compass includes an additional training phase, which remains accessible since the training phase is unconstrained (all policies are trained until convergence). Creating a COMPASS checkpoint from a pre-trained base policy involves adding parameters to process the latent vectors (\cref{app:compass-details}), resulting in a modest increase in model size. This increase never exceeds \num{2}\% of the total policy size, and has negligible impact on the overall computational or memory footprint. We report all parameter counts in~\cref{app:param-count}.

\section{Experiments}
\label{experiments}

\begin{wrapfigure}{r}{0.36\textwidth}
    \vspace{-2.0\baselineskip}
    \centering
    \includegraphics[width=\linewidth]{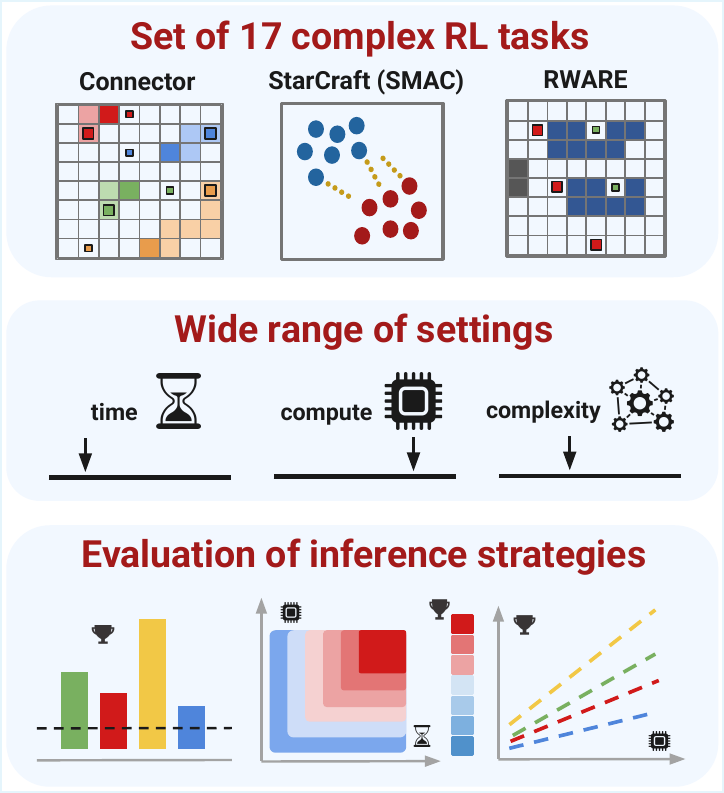}
    \caption{\textbf{Overview of our evaluation tasks and experimental study.}}
    \label{fig:experimental-overview}
    \vspace{-2.5\baselineskip}
\end{wrapfigure}

In our experimental study, we combine popular MARL algorithms and inference strategies and benchmark them on a set of complex RL tasks from the literature. Each task was specifically selected for its difficulty. We evaluate all base policies with and without inference-time search across a wide range of budget settings. Our experiments constitute the largest-ever study of inference strategies for decision-making.

\paragraph{Baselines} We use three MARL approaches to obtain our base policies: Independent PPO~\citep{dewitt2020independentlearningneedstarcraft} (\ippo) and Multi-Agent PPO~\citep{yu2022mappo} (\mappo), which are widely used and well-known MARL methods, and the recent SOTA sequence modelling approach \sable~\citep{mahjoub2025sable}. Each of these, referred to as \textit{base policies}, is evaluated with all four inference strategies introduced in~\cref{sec:main-methods}, namely stochastic sampling, \sgbs, online fine-tuning and \compass.

\paragraph{Tasks} \citet{mahjoub2025sable} established SOTA over the most comprehensive MARL benchmark published in the field to date. Interestingly, their results demonstrate that there remain certain tasks for which no existing method (including \sable) is able to achieve good performance. Specifically, these are \texttt{tiny-2ag-hard}, \texttt{tiny-4ag-hard}, \texttt{small-4ag}, \texttt{small-4ag-hard}, \texttt{medium-4ag}, \texttt{medium-4ag-hard}, \texttt{medium-6ag}, \texttt{large-4ag}, \texttt{large-4ag-hard}, \texttt{large-8ag}, \texttt{large-8ag-hard}, \texttt{xlarge-4ag} and \texttt{xlarge-4ag-hard} from Multi-Robot Warehouse~\citep{papoudakis2021benchmarking} (RWARE), \texttt{smacv2\_10\_units} and \texttt{smacv2\_20\_units} from the StarCraft Multi-Agent challenge~\citep{smac,ellis2023smacv2} (SMAC), and \texttt{con-10x10x10a} and \texttt{con-15x15x23a} from Connector~\citep{jumanji2023github}. 

\begin{wrapfigure}{r}{0.54\textwidth}
    \vspace{-1.9\baselineskip}
    \centering
    \includegraphics[width=\linewidth]{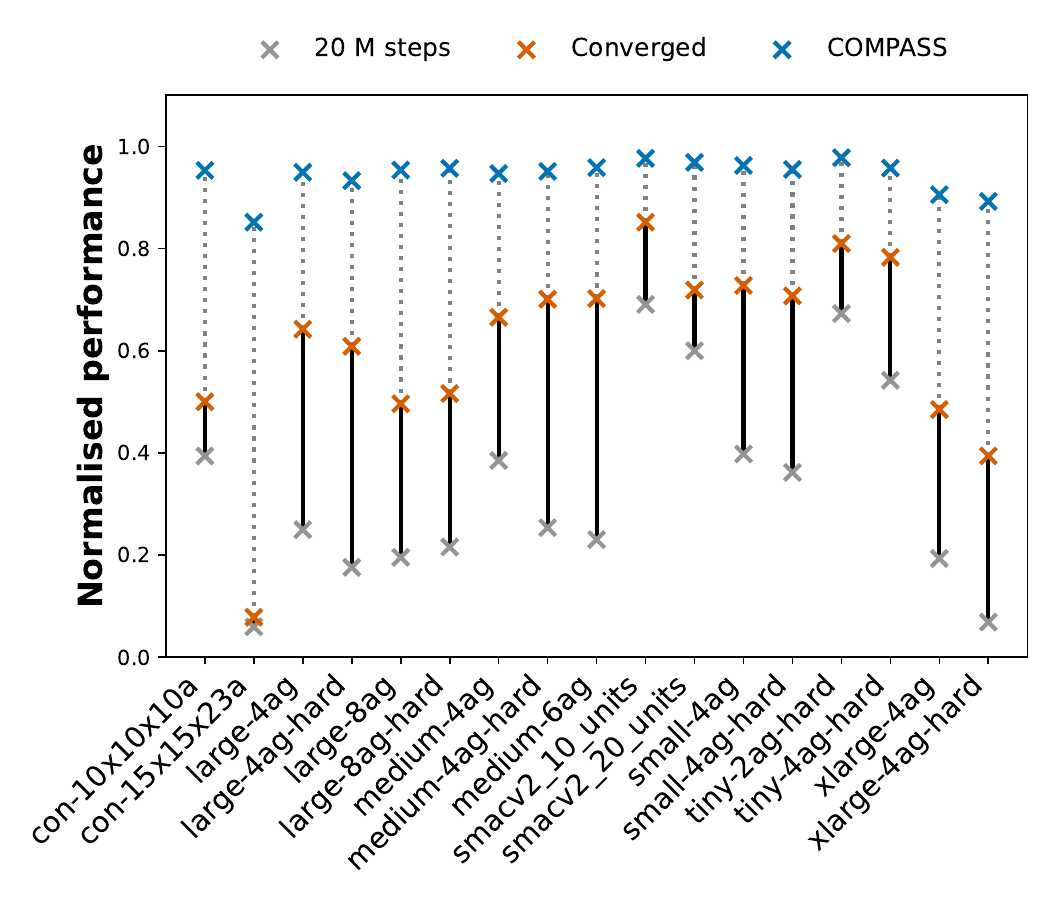}
    \caption{\textbf{Training the SOTA algorithm Sable to convergence} is not enough to achieve optimal zero-shot performance. We report the mean normalised performance per task and report 95\% bootstrap confidence intervals in~\cref{app:additional-error-bars}.}
    \label{fig:zero-shot-cvg}
    \vspace{-1.4\baselineskip}
\end{wrapfigure}

Each environment (illustrated on~\cref{fig:experimental-overview}) introduces distinct challenges that contribute to its complexity. RWARE requires agents to coordinate in order to pick up and deliver packages without collision and has a very sparse reward signal. In SMACv2 tasks, a team cooperates in real-time combat against enemies across diverse scenarios with randomised generation. Connector models the routing of a printed circuit board where agents must connect to designated targets without crossing paths. All three environments feature combinatorial and high-dimensional action spaces, partial observability and the need for tightly coordinated behaviours, making these 17 tasks a compelling test-bed for complex RL with desirable properties modelling aspects of real-world tasks. We use JAX-based implementations of Connector and RWARE from Jumanji~\citep{jumanji2023github} and for SMAC from JaxMARL~\citep{rutherford2023jaxmarl}.

\paragraph{Training base policies} To obtain clear performance ceilings for each algorithm and best isolate the effects of inference strategies, we train all base policies until convergence. For the sake of continuity with previous work with truncated training budgets, typically of \num{20}M steps, we report the zero-shot results for each converged checkpoint compared to its previous corresponding reported performance on~\cref{fig:zero-shot-cvg}. We observe that in most tasks (14 out of 17), the converged policy stays below 70\% normalised performance, demonstrating that the benchmark is still far from saturated. \compass requires an additional training phase, which reincarnates the existing base policies to create the latent space specialisation. For each base policy and task, we also train the \compass checkpoint until convergence. This leads to \num{102} trained policy checkpoints.

\paragraph{Evaluating performance during inference} Evaluating inference strategies in a way that is unbiased and aligned with real-world settings is challenging. Most papers report results where the budget is based on a number of attempts, hence not directly incorporating the time cost of the inference strategy. The time costs are reported, but it is tough to analyse due to the plurality of hardware used to obtain them. Having re-implemented all of the baselines in the same code base and setting the budget in terms of time (in seconds), we can avoid this bias in our study. We use the same fixed hardware for all our experiments, namely a \texttt{NVIDIA-A100-SXM4-80GB} GPU. For statistical robustness, we always run \num{128} independent seeds. In all cases, we control for \(\text{B}_{\text{max}}\) by varying the permitted number of batched parallel attempts instead of altering hardware between experiments. For aggregation across multiple tasks we follow the recommendations made by \citet{agarwal2021rliable} and use the \texttt{rliable} library to compute and report the inter-quartile mean (IQM) and 95\% stratified bootstrap confidence intervals.

\paragraph{Hyperparameters} To train the base policies, we re-use the hyperparameters reported in~\citet{mahjoub2025sable}, which have been optimised for our tasks. For the inference strategies, we follow recommendations from the literature~\citep{choo2022simulationguided,chalumeau2023combinatorial}. All hyperparameters choices are reported in~\cref{app:hyperparameters}.

\subsection{A couple of seconds is all you need}
\label{sec:coupleseconds}

In this section, we demonstrate that inference-time search can help reach close to maximum task performance, using base policies for which zero-shot performance stagnates around \num{60}\%.

\begin{wrapfigure}{r}{0.54\textwidth}
    \vspace{-0.2\baselineskip}
    \centering
    \includegraphics[width=\linewidth]{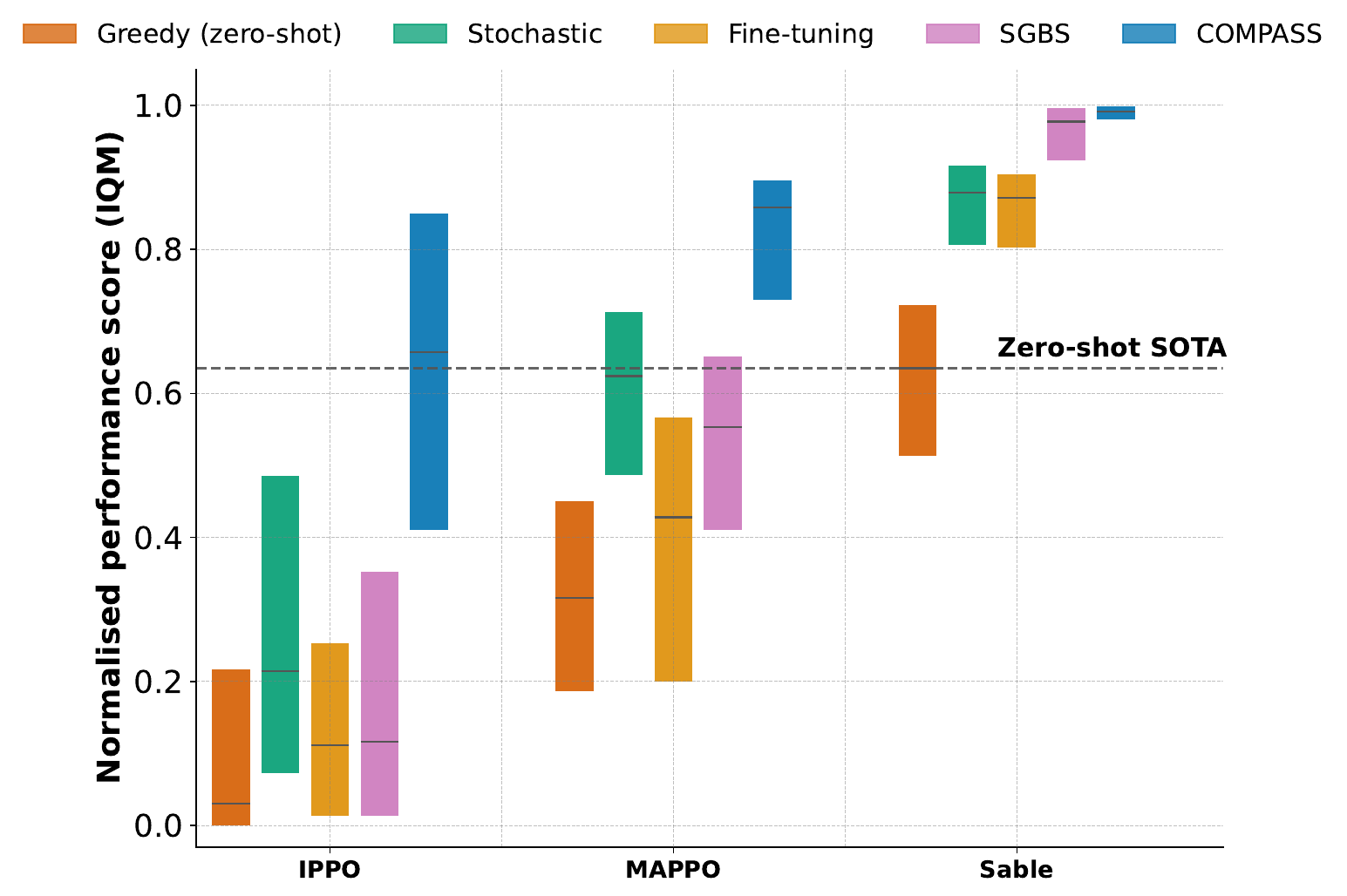}
    \caption{\textbf{Performance obtained by inference strategies over the benchmark}. Each base policy is evaluated with each possible inference strategy. We report the inter-quartile mean over tasks with 95\% stratified bootstrap confidence intervals. }
    \label{fig:inference_vs_training}
    \vspace{-0.9\baselineskip}
\end{wrapfigure}

\paragraph{Experiments} 
To demonstrate that inference-time strategies are accessible, we use a small budget: \num{30} seconds, and a compute capacity enabling to generate \num{64} solutions in parallel. Each base policy is evaluated greedily for a single attempt, and then evaluated with the search budget, using each inference strategy. We report the performance distribution over the \num{17} tasks in~\cref{fig:inference_vs_training}, and the performance gains offered by the best inference-time search over the best zero-shot on~\cref{fig:main-figure}.

\paragraph{Discussion} 
We can draw four main conclusions. First, inference-time search does provide a massive performance boost over zero-shot, which stands for every base policy. For the SOTA zero-shot method \sable, this translates to pushing the best-ever achieved aggregated performance by more than \num{45}\% and creating a system (\sable+\compass) that achieves close to \num{100}\% win-rate in all tasks where this metric is available. Second, the improvement enabled over zero-shot performance increases significantly (almost exponentially) with respect to the complexity of the task (see~\cref{fig:main-figure}). This suggests substantial gains are still ahead as the field moves toward increasingly realistic scenarios. Third, we observe that \compass is the leading strategy across tasks and base algorithms, and that \sable remains the SOTA base policy even when using inference-time search. Interestingly, under a small time budget, stochastic sampling outperforms online fine-tuning. We nevertheless show in the following section that, given more budget, this result can be nuanced, and we share our interpretation of these findings.

\subsection{Mapping performance with compute and time budget}
\label{sec:contours-exp}

A recurrent limitation in previous work on inference strategies is the use of a fixed budget during evaluation, creating a narrow view on methods, and often creating a bias towards certain types of methods. In this section, we aim at providing a much broader perspective over inference-time search by reporting performance over a grid of time and compute budgets.

\begin{figure}[t]
    \centering
    \includegraphics[width=0.98\textwidth]{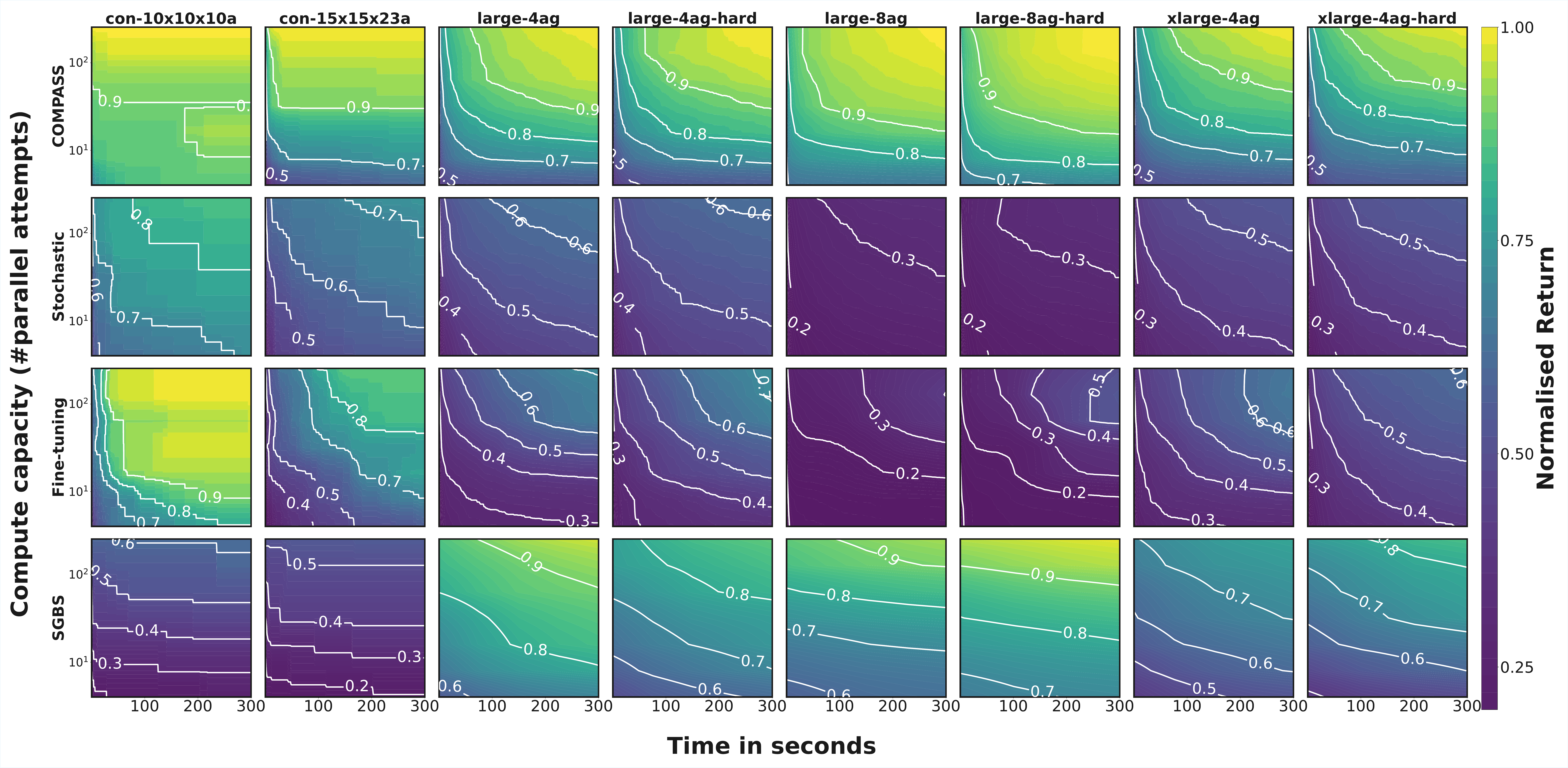}
    \caption{\textbf{Performance contour plots of Sable with different inference strategies} on a range of time budget ($x$-axis) and compute capacities ($y$-axis), for a set of representative tasks. Colours indicates performance, with brighter colours indicating higher values.}
    \label{fig:performance_contours}
\end{figure}

\paragraph{Experiments} 
We choose a maximum time of \num{300} seconds, and evaluate all inference strategies using the leading base policy (\sable), with a compute budget of $\{4, 8, 16, 32, 64, 128, 256\}$. All in all, we have \num{4} inference strategies, \num{7} possible compute budgets, \num{17} tasks, and \num{128} seeds per task, leading to \num{60928} evaluated episodes. This constitutes the largest study released on inference strategies. We report these results using contour plots, where the $x$-axis is time, $y$-axis the number of parallel attempts allowed (our proxy for compute) and colour corresponds to the performance achieved (win-rate when accessible or \textit{min}-\textit{max} normalised return) going from dark purple (\textit{min}) to yellow (\textit{max}). We keep the \num{8} hardest tasks, the lower half based on the zero-shot performance of the converged \sable checkpoints (see~\cref{fig:zero-shot-cvg}), on~\cref{fig:performance_contours} and defer remaining tasks to~\cref{app:additional-results}.

\paragraph{Discussion} 
As a sanity check, we remark that performance always increases (colours become lighter) when time or compute increases (going towards the upper right corner). We now highlight three main observations. First, \compass demonstrates impressive versatility and achieves significant gains over other inference strategies, dominating all maps, except for \texttt{con-10x10x10a}, where it gets slightly outperformed by online fine-tuning. Second, we observe high variance for online fine-tuning: getting close to \compass for large budgets on \texttt{con-10x10x10a}, yet struggling to match stochastic sampling on others (e.g., RWARE's \texttt{large-8ag}). This shows that fine-tuning can be detrimental by reducing the number of attempts made within the time budget. Plus, policy gradients can be unstable (small batch size) or converge to local optima. This observation disproves the common belief that inference-time search is as trivial as over-fitting to the problem instance. Finally, we observe that \sgbs, despite failing on the connector tasks, achieves competitive performance overall and even leads in the low-budget regime (i.e., below 100 seconds and fewer than 10 parallel attempts) in some scenarios (\texttt{large-4ag} and \texttt{large-4ag-hard}).

\subsection{Scaling with increasing budget}
\label{sec:scaling-budget}

In practical applications, time is often more restricted than compute: a couple of seconds or minutes can be allowed, sometimes a few hours (train scheduling for the next day), but rarely more. Being able to improve solutions' quality under a fixed time budget by increasing compute is desirable. In this section, we analyse how different methods scale with additional compute.

\begin{wrapfigure}{r}{0.53\textwidth}
    \vspace{-2.5\baselineskip}
    \centering
    \includegraphics[width=\linewidth]{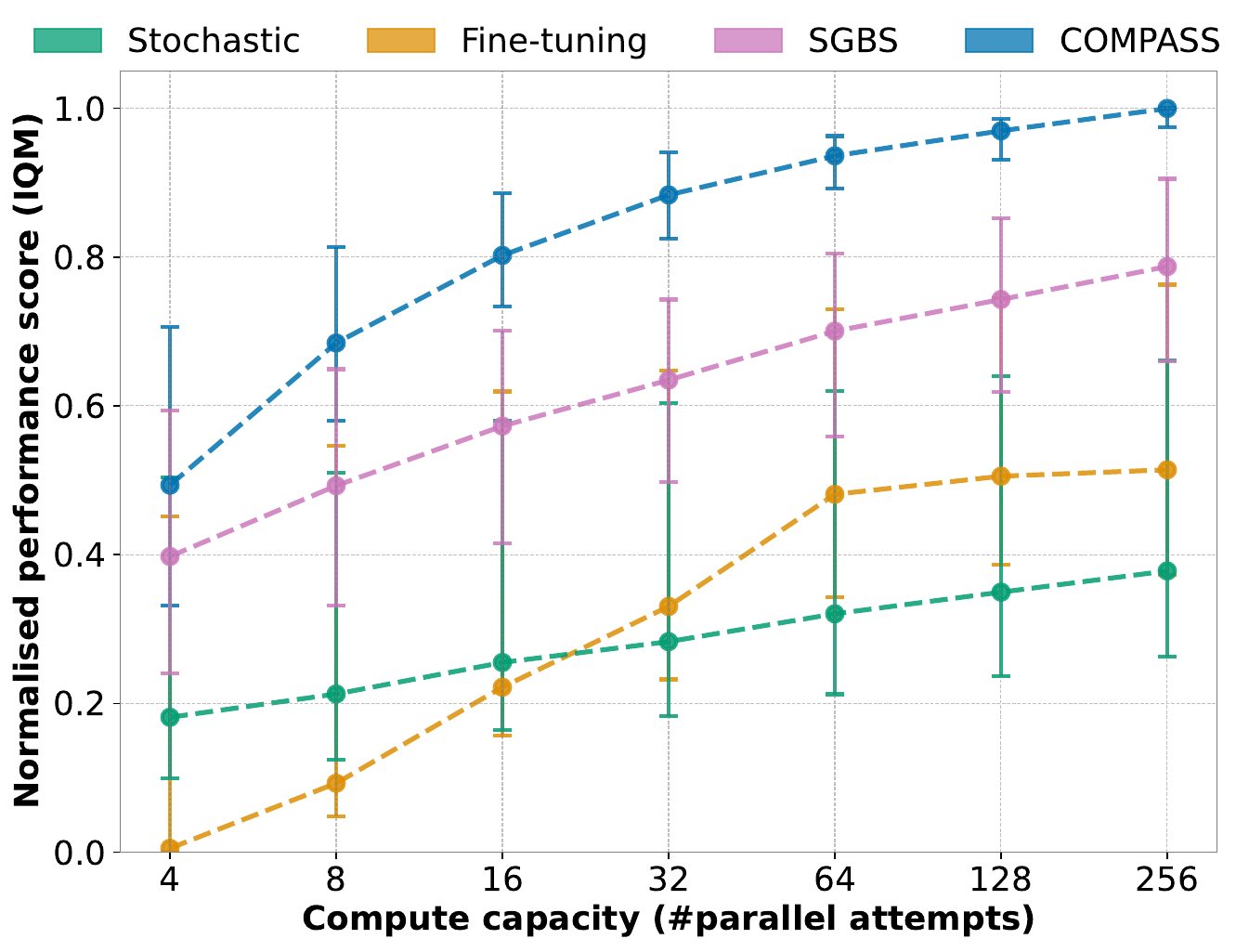}
    \caption{\textbf{Scaling of performance with respect to compute capacity (fixed time budget).} Compute capacity is set by limiting the number of parallel attempts. \sable+\compass demonstrates impressive scaling properties, reaching the performance bounds.}
    \label{fig:scaling_trends}
    \vspace{-2.5\baselineskip}
\end{wrapfigure}

\paragraph{Experiments} 
We keep the time budget fixed, \num{300} seconds, and we plot the final performance for each possible compute budget (still using the number of parallel attempts as a proxy). We use \sable as the base policy and evaluate all the inference strategies across the \num{8} hardest tasks. We report results per strategy, and aggregated over the tasks on~\cref{fig:scaling_trends}.

\paragraph{Discussion} 
As expected, stochastic sampling has the lowest scaling coefficient: its search solely relies on chance, no adaptation or additional search is happening. On the other hand, we can see that online fine-tuning benefits from more compute, probably due to a better estimation of the policy gradient. It nevertheless requires a budget of 64 parallel attempts to clearly outperform stochastic sampling. On the other hand, \compass consistently provides a significant advantage. Its scaling trend seems linear at first, with a high coefficient, and only seems to decline for higher budgets, as performance limits are reached (over \num{95}\% win-rate when accessible, best-ever observed performance elsewhere). We have a two-fold explanation for these particularly impressive scaling properties: (i) the diversity contained in the latent space can be exploited by more parallelism, leading to a massive exploration of the solution space, even when the initial policy is far from optimal, and (ii) the higher the batch size, the better the \cmaes search, enabling \compass to exploit even more information from any additional searching step allowed within the given time budget. We discuss this further in~\cref{app:why-compass-better}.

\section{Conclusion}

In this work, we demonstrate that inference-time strategies are a critical and underutilized lever for boosting the performance of RL systems in complex tasks, using multi-agent RL as a representative test-bed. While training-time improvements have long dominated the field, our results show that inference-time search may offer significant performance gains using only a few seconds to minutes of additional wall-clock time during execution. We introduce a unified view of inference strategies, extend it to the multi-agent setting, and empirically validate its effectiveness under varying compute and time budgets.

Our large-scale evaluation, the most comprehensive to date on inference-time methods, reveals three \textbf{key takeaways}: (i) inference-time search with a relevant strategy yields significant improvements, even under tight time constraints; (ii) the gains depend on the inference budget, and our contour maps provide practitioners with practical guidance based on their constraints; and (iii) \sable+\compass not only dominates the benchmark but also exhibits the most favourable scaling trends, making it particularly effective for increasingly complex decision-making problems.

Altogether, our findings call for a shift in how decision-making models are evaluated and deployed: inference-time strategies should be treated as core components of the solution pipeline, not as optional refinements. We hope our results and open-source tools will encourage broader adoption and inspire further innovation in the design of scalable inference-time algorithms.

\paragraph{Limitations and future work} We focus on multi-agent RL as it better captures the complexity of real-world decision-making systems, where successful operation often relies on coordination among multiple agents. Nonetheless, we acknowledge that naturally single-agent tasks can also be highly complex. In the spirit of seeding a wider investigation for future work beyond the multi-agent setting, we provide additional results on the single-agent Craftax benchmark in~\cref{app:single-agent}, showing that a simple \num{30}-second inference-time search using stochastic sampling can provide a \num{37}\% performance boost. This provides initial (but limited) evidence that our claims hold more generally and we leave a more thorough investigation for future work. Beyond the broader RL setting, we intend to investigate two main future research directions. First, studying how to best combine existing inference paradigms to leverage their complementary strengths. Second, investigating how inference strategies compare when evaluated out-of-distribution.


\section*{Acknowledgements}
We would like to thank Guillaume Toujas-Bernate, Jake Lourie and Thomas Lecat for useful discussions on the use of inference strategies in real-world applications. We thank our MLOps
team for developing our model training and experiment orchestration platform \href{https://aichor.ai/}{AIchor}. We thank the Python and JAX communities for developing tools that made this research possible. We thank the anonymous reviewers for their constructive feedback and valuable suggestions. Finally, we thank Google's TPU Research Cloud (TRC) for supporting our research with Cloud TPUs.

\newpage

\bibliographystyle{abbrvnat}
\bibliography{references}




\newpage
\appendix
\section*{Appendix}

\section{Additional results}
\label{app:additional-results}
\begin{figure}[h]
    \centering
    \includegraphics[width=0.98\textwidth]{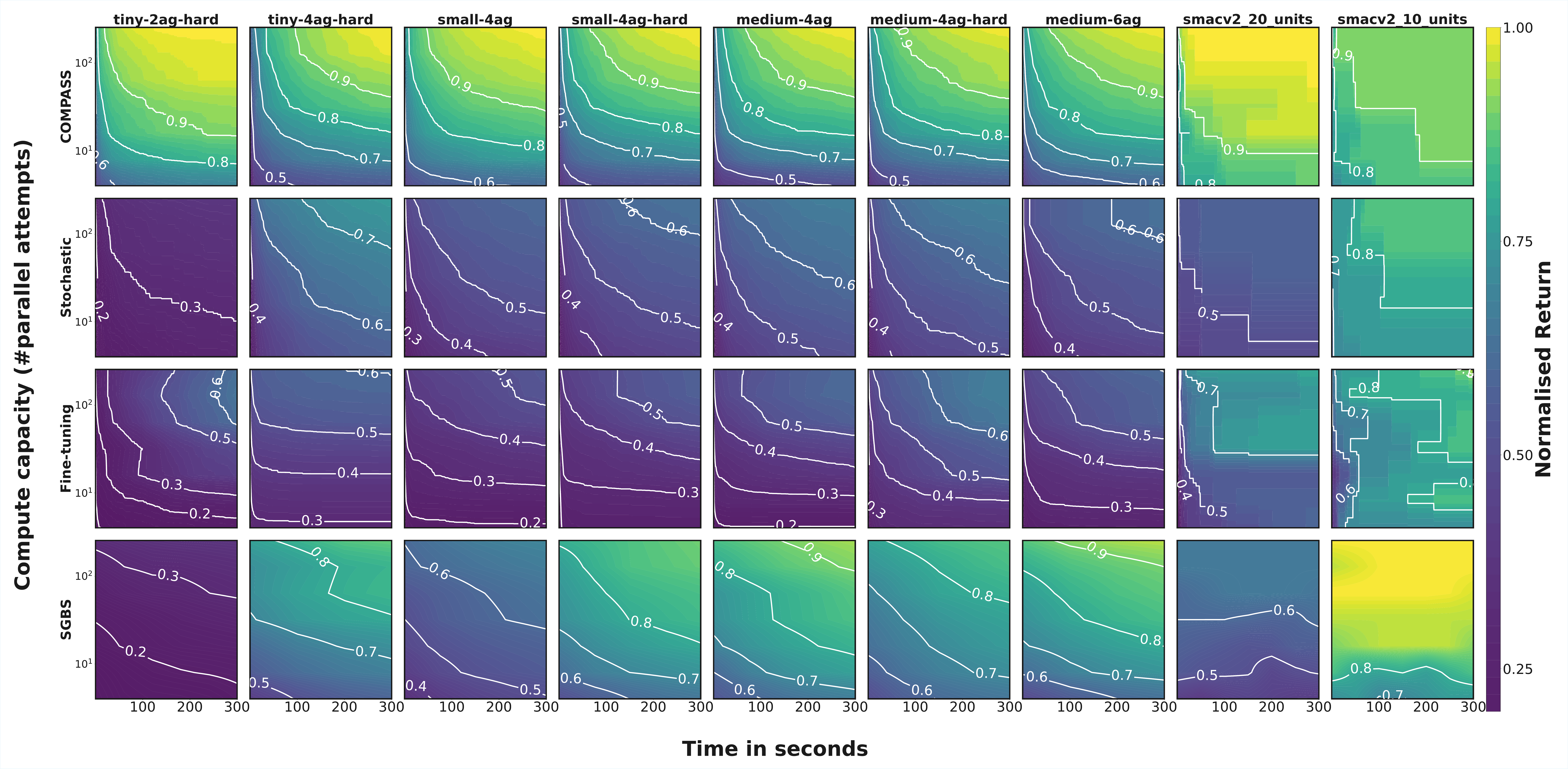}
    \caption{\textbf{Performance contour plots of Sable with different inference strategies on remaining 9 tasks} on a range of time budget ($x$-axis) and compute capacities ($y$-axis). Colours indicates performance, with brighter colours indicating higher values.}
    \label{fig:performance_contours_remaining}
\end{figure}

\cref{sec:contours-exp} presents the contour plots of Sable with all inference strategies on the hardest 8 tasks of the benchmark. We report the contour plots of the other \num{9} remaining tasks on~\cref{fig:performance_contours_remaining}.

\cref{fig:performance_contours_remaining}, shows similar trends to those illustrated in~\cref{sec:contours-exp}. \sable+\compass leads to the best overall performance. Stochastic sampling provides a good and robust baseline overall. Online fine-tuning suffer from variance: despite being able to compete close to \compass on a couple tasks, it often get outperformed by stochastic sampling. Finally, \sgbs obtains good performance on numerous scenarios, and even leads one scenario (\texttt{smacv2\_10\_units}).

\section{Details about the tasks}
\label{app:tasks-details}

\begin{figure}[h!]
    \centering
    \includegraphics[height=5cm,keepaspectratio]{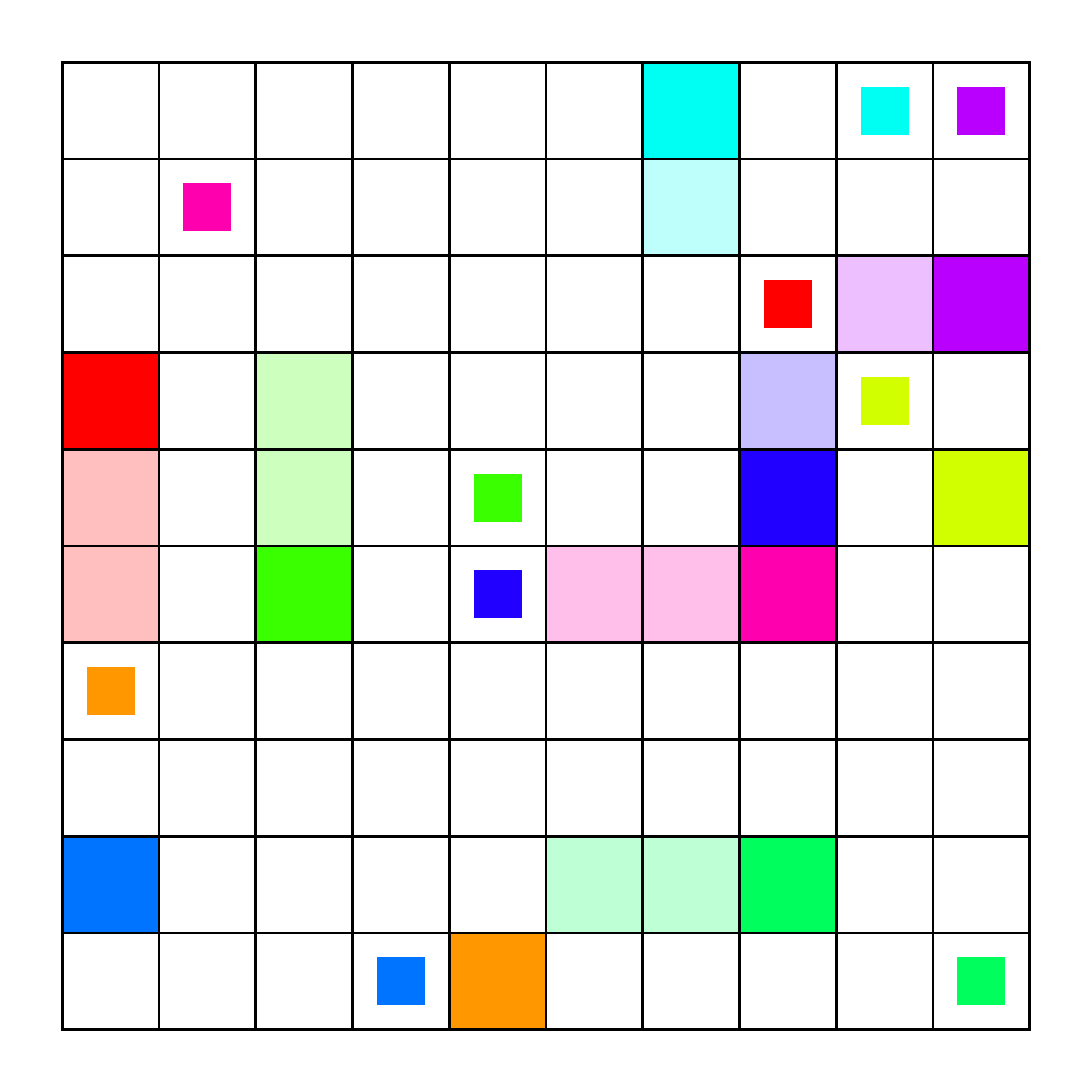}
    \caption{Environment rendering for Connector. Task name: \texttt{con-10x10-10a}. Image from~\citet{mahjoub2025sable}.}
    \label{fig:connector-task}
\end{figure}

\paragraph{Connector} is an environment designed to model the routing of a printed circuit board (PCB). Agents start at randomly generated initial positions and have to reach a goal position without overlapping with each other's paths (indicated as lower opacity coloured cells in Figure~\ref{fig:connector-task}). Each agent has a partial view with a fixed field of view around itself as well as its current $(x, y)$-coordinate on the grid and the $(x, y)$-coordinate of its goal location. At each timestep, agents receive a small negative reward $-0.03$ and receive a reward of $1$ for successfully connecting to a goal location. The particular difficulty in this environment stems from the fact that agents have to select actions carefully so that they not only reach their goal greedily but so that all agents can ultimately reach their goals. 

\begin{figure}[h!]
    \centering
    \includegraphics[height=5cm,keepaspectratio]{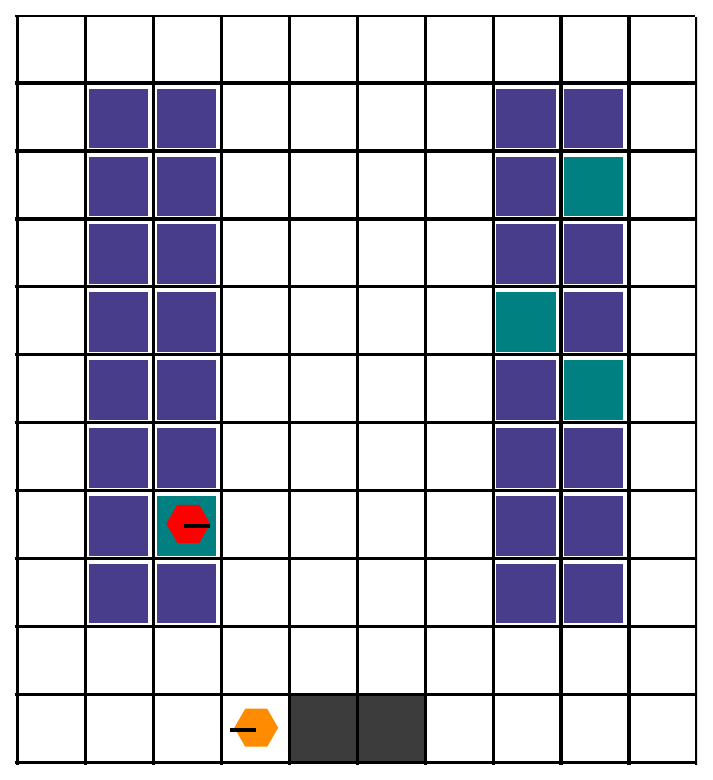}
    \caption{Environment rendering for Robot Warehouse. Task name: \texttt{tiny-2ag}. Image from~\citet{mahjoub2025sable}.}
    \label{fig:rware}
\end{figure}

\paragraph{RWARE} is an environment where agents need to coordinate to deliver shelves (green cells in Figure \ref{fig:rware}) to a goal location (dark grey cells in Figure \ref{fig:rware}) in a warehouse. The reward is sparse since agents only receiving a reward of $1$ for a successful shelf delivery to the goal location and $0$ otherwise. This sparsity makes the environment particularly difficult since agents have to complete a long sequence of correct actions in order to receive any reward signal.

\begin{figure}[h!]
    \centering
    \includegraphics[height=5cm,keepaspectratio]{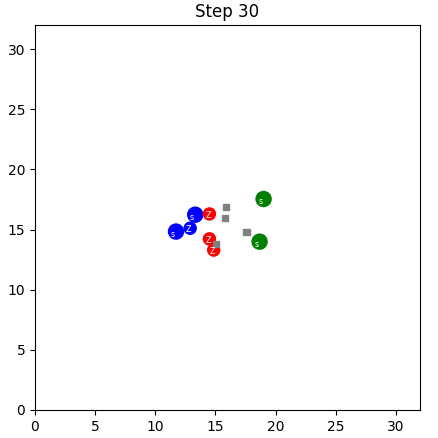}
    \caption{Environment rendering for SMAX. Task name: \texttt{2s3z}. Image from~\citet{mahjoub2025sable}.}
    \label{fig:smax}
\end{figure}

\paragraph{SMAC} is an environment where agents need to coordinate and develop the correct micro strategies in order to defeat an enemy team. It was shown that the original benchmark was overly simplistic and for this reason SMACv2~\citep{ellis2023smacv2} was developed in order to address the shortcomings of the original benchmark. SMACv2 randomly generates a team of allied units by sampling from a selection of unit types and randomly generates ally starting positions at the start of each episode. The need for generalisation is what gives this benchmark its difficulty.

\paragraph{JAX-based implementations.} Since Connector was developed as part of Jumanji \citep{jumanji2023github}, we make use of the JAX-based implementation directly. For RWARE we use the JAX-based reimplementation from Jumanji and for SMAC and SMACv2 we make use of the JAX-based reimplementation from JaxMARL \citep{rutherford2023jaxmarl}, named SMAX. For a detailed discussion on task naming conventions we refer the reader to the Appendix of~\citet{mahjoub2025sable}.

\section{Results and discussion about the single-agent case}
\label{app:single-agent}

\subsection{Additional results on a single-agent challenge: Craftax}

To demonstrate that our core message is not limited to the multi-agent case, we selected a single-agent task that is still far from being solved by RL algorithms, Craftax~\citep{matthews2024craftax}.  

We trained a \sable base policy and matched the current state-of-the-art performance reported on the Craftax-1B leaderboard, which is \num{18.3}\% of the maximum score. We then ran a \num{30}-second inference-time search with the simplest inference strategy (stochastic sampling). The results are reported in~\cref{tab:craftax-results}. Strikingly, and similar to our main results, using only a \num{30}-second inference-time search yields a \num{37}\% performance gain, reaching \num{25.8}\% of the maximum score. To further put this into context, this performance boost is of the same magnitude as that of the last \num{300} million steps of training the base policy.

These new results confirm that the significant improvement enabled by inference-time search is not limited to multi-agent RL and is applicable to RL in general. In the next subsection, we discuss existing results (reported in the literature) to further support the importance of inference-time search in single-agent settings.

\begin{table}[h!]
    \centering
    \caption{Performance comparison on Craftax-1B between zero-shot inference and stochastic sampling with a 30-second budget. Values indicate mean performance with 95\% confidence intervals.}
    \label{tab:craftax-results}
    \begin{tabular}{lcc}
    \toprule
    \textbf{Scenario} & \textbf{Zero-shot} & \textbf{Stochastic Sampling (30s budget)} \\
    \midrule
    Craftax-1B & 18.8 (18.1, 19.5) & 25.8 (25.7, 26.1) \\
    \bottomrule
    \end{tabular}
\end{table}

\subsection{Published results on single-agent tasks}

To further demonstrate the impact of inference-time strategies beyond the multi-agent case, we discuss previous results reported in the literature. These results do not show the same impact because they are often close to saturation, even with zero-shot. They nevertheless demonstrate that inference-time strategies can be successfully applied to a wide range of problems and provide noticeable improvement.

We refer to the experimental results reported in the seminal works that introduced the inference strategies used in our paper~\citep{chalumeau2023combinatorial,choo2022simulationguided,Bello16}. All of these evaluate on single-agent tasks (TSP, CVRP, JSSP, Knapsack, FFSP) and show that inference strategies bring statistically significant improvements over zero-shot performance. 

However, we recall the limitations we have identified on these experimental results: the limited breadth of the time/compute settings, methods’ reliance on problem-specific tricks and saturated benchmarks where zero-shot performance is already very high.

\section{Hyperparameters}
\label{app:hyperparameters}

In this section, we report the hyper-parameters used in our experiments. They can also be found in the submitted code files.

\paragraph{Training - base policies} Hyper-parameters used to train the base policies for this benchmark are taken from~\citet{mahjoub2025sable}. These hyper-parameters were tuned on each task with a budget of 40 trials using the Tree-structured Parzen Estimator (TPE) Bayesian optimisation
algorithm, and are reported in~\citet{mahjoub2025sable}.

\paragraph{Hyper-parameters - COMPASS Training} Hyper-parameters used to train all the \compass checkpoints can be found in~\cref{tab:compass-training}. Most of them were kept close to the original hyper-parameters used in~\citet{chalumeau2023combinatorial}.

\begin{table}[h]
\centering
\caption{Hyper-parameters used for \compass' training phase. The same values are used for all three base policies (i.e., \sable, \ippo, \mappo).}
\label{tab:compass-training}
\begin{tabular}{ll}
\toprule
\textbf{Parameter} & \textbf{Value} \\
\midrule
Instances batch size & 128 \\
Latent space sample size & 64 \\
Latent space dimension size & 16 \\
Latent amplifier & 1 \\
Padding with random weights & True \\
Weight noise amplitude & 0.01 \\
\bottomrule
\end{tabular}
\end{table}

\paragraph{Hyper-parameters - Inference-time search} Most values are directly taken from original works or kept close to these. When different, we tried to use guidance from original work to decide how to set them. The values for the \cmaes search used with \compass, for all base policies, can be found in~\cref{tab:cmaes-config}. The values for online fine-tuning (all base policies) are reported in~\cref{tab:finetuning-config}. Hyper-parameters for \sgbs (all base policies) are reported in~\cref{tab:sgbs-config}, and for stochastic sampling in~\cref{tab:Stochastic-config}.

\begin{table}[h]
\centering
\caption{Hyper-parameters of the \cmaes process used to search \compass' latent space at inference time. The same parameters are used for all tasks and all base policies (i.e., \sable, \mappo and \ippo).}
\label{tab:cmaes-config}
\begin{tabular}{lccccccc}
\toprule
\textbf{Parameter} & \multicolumn{7}{c}{\textbf{Number of attempts}} \\
\cmidrule(lr){2-8}
& \textbf{4} & \textbf{8} & \textbf{16} & \textbf{32} & \textbf{64} & \textbf{128} & \textbf{256} \\
\midrule
Latent sample size & 4 & 8 & 16 & 32 & 64 & 128 & 256 \\
Number of elites & 2 & 4 & 8 & 16 & 32 & 64 & 128 \\
Covariance matrix step size & 1.0 & 1.0 & 1.0 & 1.0 & 1.0 & 1.0 & 1.0 \\
Initial sigma & 1.0 & 1.0 & 1.0 & 1.0 & 1.0 & 1.0 & 1.0 \\
Num. components & 1 & 1 & 1 & 1 & 1 & 1 & 1 \\
\bottomrule
\end{tabular}
\end{table}

\begin{table}[H]
\centering
\caption{Hyper-parameters used when doing online fine-tuning at inference-time with \sable, \mappo and \ippo.}
\label{tab:finetuning-config}
\begin{tabular}{llccc}
\toprule
\textbf{Parameter} & \textbf{Task Name} & \textbf{Sable} & \textbf{IPPO} & \textbf{MAPPO} \\
\midrule
\multirow{1}{*}{Max. trajectory size} 
& All tasks & 64 & 64 & 64 \\
\midrule
\multirow{17}{*}{Learning rate} 
& \texttt{con-10x10x10a}       & 0.001   & 0.00025  & 0.0001  \\
& \texttt{con-15x15x23a}       & 0.001   & 0.0001   & 0.0001  \\
& \texttt{large-4ag}           & 0.001   & 0.0005   & 0.0001  \\
& \texttt{large-4ag-hard}      & 0.001   & 0.0005   & 0.00025 \\
& \texttt{large-8ag}           & 0.001   & 0.00025  & 0.0001  \\
& \texttt{large-8ag-hard}      & 0.001   & 0.00025  & 0.0005  \\
& \texttt{medium-4ag}          & 0.001   & 0.00025  & 0.00025 \\
& \texttt{medium-4ag-hard}     & 0.0005  & 0.0005   & 0.00025 \\
& \texttt{medium-6ag}          & 0.0005  & 0.0001   & 0.00025 \\
& \texttt{smacv2\_10\_units}   & 0.00025 & 0.0005   & 0.00025 \\
& \texttt{smacv2\_20\_units}   & 0.001   & 0.0005   & 0.0005  \\
& \texttt{smacv2\_5\_units}    & 0.001   & 0.00025  & 0.00025 \\
& \texttt{small-4ag}           & 0.0005  & 0.0005   & 0.0005  \\
& \texttt{small-4ag-hard}      & 0.001   & 0.0001   & 0.00025 \\
& \texttt{tiny-2ag-hard}       & 0.001   & 0.00025  & 0.00025 \\
& \texttt{tiny-4ag-hard}       & 0.0005  & 0.0005   & 0.0005  \\
& \texttt{xlarge-4ag}          & 0.001   & 0.0005   & 0.0005  \\
& \texttt{xlarge-4ag-hard}     & 0.0001  & 0.0005   & 0.0001  \\
\midrule
\multirow{3}{*}{Entropy coefficient} 
& Connector & 0.05 & 0.00 & 0.00  \\
& RWARE & 0.0 & 0.0 & 0.0  \\
& SMAX & 0.0 & 0.0 & 0.0  \\
\bottomrule
\end{tabular}
\end{table}

\begin{table}[H]
\centering
\caption{Hyper-parameters for Simulation Guided Beam Search (\sgbs). The same values are used for \sable, \ippo and \mappo. Values are chosen such that expansion factor and beam width (top-k) use fully the number of attempts allowed by batch.}
\label{tab:sgbs-config}
\begin{tabular}{lcccccc}
\toprule
\textbf{Parameter} & \multicolumn{6}{c}{\textbf{Number of attempts}} \\
& \textbf{4} & \textbf{16} & \textbf{32} & \textbf{64} & \textbf{128} & \textbf{256} \\
\midrule
Expansion factor & 4 & 4 & 4 & 4 & 4 & 4 \\
Beam width (Top k) & 1 & 4 & 8 & 16 & 32 & 64 \\
\bottomrule
\end{tabular}
\end{table}

\begin{table}[H]
\centering
\caption{To sample stochastically from a policy (\sable, \ippo or \mappo), we always use the same temperature as the one used during training: \num{1.0}.}
\label{tab:Stochastic-config}
\begin{tabular}{ll}
\toprule
\textbf{Parameter} & \textbf{Value} \\
\midrule
Evaluation greedy & False \\
Temperature & 1.0 \\
\bottomrule
\end{tabular}
\end{table}

\section{Experimental stack}

All algorithmic implementations are built by extending the JAX-based research library, \texttt{Mava}~\citep{de2021mava}. Our version of \cmaes used for searching \compass's latent space comes from the JAX-based quality diversity library, \texttt{QDAX}~\citep{chalumeau2024qdax}. Both these libraries are built to leverage the DeepMind JAX ecosystem \citep{deepmind2020jax} and use \texttt{Flax} \citep{flax2020github} for building neural networks, \texttt{optax} for optimisers and \texttt{orbax} for model checkpointing.

\section{Additional details about training}

\subsection{Base policy training}

In~\cref{method}, we explain that in our setting, it is assumed that all methods can be trained until convergence. For transparency and comparison with previous results in the literature, we report in ~\cref{tab:training_budgets} the training budget given to the methods to reach convergence, in environment steps.

\begin{table}[H]
\centering
\caption{Training budgets (in environment steps) for each method across tasks.}
\label{tab:training_budgets}
\begin{tabular}{lccc}
\toprule
\textbf{Task Name} & \textbf{Sable} & \textbf{MAPPO} & \textbf{IPPO} \\
\midrule
\texttt{con-10x10x10a}     & 100M & 200M & 200M \\
\texttt{con-15x15x23a}     & 100M & 200M & 200M \\
\texttt{large-4ag}         & 600M & 200M & 200M \\
\texttt{large-4ag-hard}    & 600M & 400M & 200M \\
\texttt{large-8ag}         & 600M & 400M & 200M \\
\texttt{large-8ag-hard}    & 600M & 200M & 200M \\
\texttt{medium-4ag}        & 400M & 200M & 200M \\
\texttt{medium-4ag-hard}   & 600M & 200M & 200M \\
\texttt{medium-6ag}        & 600M & 200M & 200M \\
\texttt{small-4ag}         & 200M & 200M & 200M \\
\texttt{small-4ag-hard}    & 400M & 200M & 200M \\
\texttt{tiny-2ag-hard}     & 100M & 200M & 200M \\
\texttt{tiny-4ag-hard}     & 200M & 200M & 200M \\
\texttt{xlarge-4ag}        & 200M & 400M & 200M \\
\texttt{xlarge-4ag-hard}   & 200M & 200M & 200M \\
\texttt{smacv2\_10\_units}  & 100M & 200M & 200M \\
\texttt{smacv2\_20\_units}  & 600M & 200M & 200M \\
\bottomrule
\end{tabular}
\end{table}

\subsection{COMPASS training}
\label{app:compass-details}

In this section, we provide details about \compass and our implementation. For a more in-depth explanation, we refer the reader to the original paper~\citep{chalumeau2023combinatorial}. Our implementation is available in the submitted code files.

\compass is a method, which (i) uses a pre-trained base policy and modifies it to be conditioned by a latent vector (ii) re-trains it to enable latent vectors to create policies that are diversified and specialised for distinct training sub-distributions. At inference-time, this enables to search for the latent vector which performs best on the new instance to be solved. 

\paragraph{Reincarnating a pre-trained base policy} \compass checkpoints are not trained from scratch, they reincarnate an existing base policy, add parameters to this policy in order to process the latent vector which can be given as additional input. Hence, one must choose where to inject the latent vector in the existing architecture, and must modify the existing parameters of the neural policy to account for the new shape of the input. The latent vector must be input in a way that enables to diversify the actions produced for the same observation (i.e., making sure that different latent vectors create different policies). In practice, we concatenate the latent vector to the observations. When relevant, we make sure that this is done after observation normalisation layers. 

In the original work, the additional weights added to the base neural network are initialised with zeros. In our case, we observed that this could not provide enough diversification in the first training step, preventing any emergence of specialisation, leading to no benefit. We found that initialising these parameters with random uniform values between \num{-0.01} and \num{0.01} fixed the issue across all tasks.

\paragraph{Creating diversity and specialisation in the latent space} Similarly to the original method, the latent space is not learned, it is a fixed prior, and the network's weights are learned to create diversity from this prior space. We also use a similar prior, which is a uniform distribution between \num{-1} and \num{1}, over \num{16} dimensions. We do not need any amplification of the latent vector (i.e., we multiply by \num{1}, whereas the original work multiplies it by \num{100}). At each training step, a batch of instances is sampled from the training distribution, a batch of latent vectors is sampled from that latent space, the conditioned policy is evaluated for each instance, for each latent vector, and only the best performing latent vectors, for each instance, are used in the computation of the loss. This creates the specialisation (diversification). For additional motivation and mathematical formulation, we refer the reader to~\citet{chalumeau2023combinatorial}. Like all training processes in our experimental study, the method is allowed to train until convergence. In practice, each \compass checkpoint was trained for \num{100} million steps.

\section{Pseudo-algorithms}

In this section, we provide algorithmic descriptions for the inference strategies used in the paper, showing how the policy is used at inference time, under the time and budget constraints. Stochastic sampling is explained in~\cref{alg:stochastic-sampling}, \sgbs in~\cref{alg:sgbs}, online fine-tuning in~\cref{alg:online-finetuning} and \compass in~\cref{alg:compass}.

Algorithms use the terminology \textit{trajectory} $\tau$, which is equivalent to \textit{solution}. The symbol $\circ$ is used for the concatenation of a new step transition to a partial trajectory.

\begin{algorithm}
    \caption{Stochastic Sampling}\label{alg:stochastic-sampling}
    \begin{algorithmic}[1]
        \STATE \textbf{Input:} policy $\boldsymbol{\pi}_\theta$, instance $\rho$, time budget $T_{\max}$, compute capacity $B_{\max}$, reward function $\mathcal{R}$
        \STATE Initialize best trajectory $\tau^{*} \leftarrow \emptyset$, best score $R^{*} \leftarrow -\infty$
        \WHILE{elapsed time $< T_{\max}$}
            \FOR{parallel rollout $b = 1$ to $B_{\max}$}
                \STATE Initialize trajectory $\tau_b \leftarrow \emptyset$
                \FOR{step $t = 1$ to $H$}
                    \STATE Observe $\boldsymbol{o}_t$ from environment copy $\rho_b$
                    \STATE Sample $\boldsymbol{a}_t \sim \boldsymbol{\pi}_\theta(\cdot \mid \boldsymbol{o}_t)$
                    \STATE Execute $\boldsymbol{a}_t$ in $\rho_b$, observe $\boldsymbol{o}_{t+1}$
                    \STATE Append $(\boldsymbol{o}_t, \boldsymbol{a}_t)$ to $\tau_b$
                \ENDFOR
                \STATE $R_b \leftarrow \mathcal{R}(\tau_b)$
                \IF{$R_b > R^{*}$}
                    \STATE $\tau^{*} \leftarrow \tau_b$, \quad $R^{*} \leftarrow R_b$
                \ENDIF
            \ENDFOR
        \ENDWHILE
        \STATE \textbf{return} $\tau^{*}$
    \end{algorithmic}
\end{algorithm}

\begin{algorithm}
    \caption{Simulation-Guided Beam Search}\label{alg:sgbs}
    \begin{algorithmic}[1]
        \STATE \textbf{Input:} policy $\boldsymbol{\pi}_\theta$, instance $\rho$, time budget $T_{\max}$, compute capacity $B_{\max}$, beam width $K$, reward function $\mathcal{R}$
        \STATE Initialize beam $\mathcal{B} \leftarrow \{\emptyset\}$, best trajectory $\tau^{*} \leftarrow \emptyset$, best score $R^{*} \leftarrow -\infty$
        \WHILE{elapsed time $< T_{\max}$}
            \STATE $\mathcal{B}_{\text{new}} \leftarrow \emptyset$
            \FOR{partial trajectory $\tau$ in $\mathcal{B}$}
                \FOR{sample $b = 1$ to $B_{\max} / |\mathcal{B}|$}
                    \STATE Let $\boldsymbol{o}_t$ be the observation at the end of $\tau$
                    \STATE Sample $\boldsymbol{a}_t \sim \boldsymbol{\pi}_\theta(\cdot \mid \boldsymbol{o}_t)$
                    \STATE Simulate greedy rollout from $\tau \circ \boldsymbol{a}_t$ using $\boldsymbol{\pi}_\theta$
                    \STATE Let $\hat{\tau}_b$ be the resulting full trajectory
                    \STATE $R_b \leftarrow \mathcal{R}(\hat{\tau}_b)$
                    \STATE Add $(\tau \circ \boldsymbol{a}_t, R_b)$ to $\mathcal{B}_{\text{new}}$
                    \IF{$R_b > R^{*}$}
                        \STATE $\tau^{*} \leftarrow \hat{\tau}_b$, \quad $R^{*} \leftarrow R_b$
                    \ENDIF
                \ENDFOR
            \ENDFOR
            \STATE Prune top-$K$ trajectories from $\mathcal{B}_{\text{new}}$ into $\mathcal{B}$
        \ENDWHILE
        \STATE \textbf{return} $\tau^{*}$
    \end{algorithmic}
\end{algorithm}

\begin{algorithm}
    \caption{Online Fine-Tuning}\label{alg:online-finetuning}
    \begin{algorithmic}[1]
        \STATE \textbf{Input:} base policy $\boldsymbol{\pi}_\theta$, instance $\rho$, time budget $T_{\max}$, compute capacity $B_{\max}$, reward function $\mathcal{R}$, learning rate $\alpha$
        \STATE Initialize best trajectory $\tau^{*} \leftarrow \emptyset$, best score $R^{*} \leftarrow -\infty$
        \STATE Initialize adapted parameters $\theta' \leftarrow \theta$
        \WHILE{elapsed time $< T_{\max}$}
            \FOR{parallel rollout $b = 1$ to $B_{\max}$}
                \STATE Rollout $\tau_b \sim \boldsymbol{\pi}_{\theta'}$ on $\rho$
                \STATE $R_b \leftarrow \mathcal{R}(\tau_b)$
                \IF{$R_b > R^{*}$}
                    \STATE $\tau^{*} \leftarrow \tau_b$, \quad $R^{*} \leftarrow R_b$
                \ENDIF
            \ENDFOR
            \STATE Compute gradient: $\nabla_\theta \leftarrow \frac{1}{B_{\max}} \sum_{b=1}^{B_{\max}} \nabla_\theta \log \boldsymbol{\pi}_{\theta'}(\tau_b) \cdot R_b$
            \STATE Update parameters: $\theta' \leftarrow \theta' + \alpha \nabla_\theta$
        \ENDWHILE
        \STATE \textbf{return} $\tau^{*}$
    \end{algorithmic}
\end{algorithm}

\begin{algorithm}
    \caption{COMPASS + CMA-ES search}\label{alg:compass}
    \begin{algorithmic}[1]
        \STATE \textbf{Input:} latent-conditioned policy $\boldsymbol{\pi}_\theta(\cdot \mid \boldsymbol{o}, z)$, instance $\rho$, time budget $T_{\max}$, compute capacity $B_{\max}$, reward function $\mathcal{R}$
        \STATE Initialize CMA-ES: mean $\mu$, covariance $\Sigma$
        \STATE Initialize best trajectory $\tau^{*} \leftarrow \emptyset$, best score $R^{*} \leftarrow -\infty$
        \WHILE{elapsed time $< T_{\max}$}
            \STATE Sample $\{z_b\}_{b=1}^{B_{\max}} \sim \mathcal{N}(\mu, \Sigma)$
            \FOR{each $z_b$ in parallel}
                \STATE Rollout $\tau_b \sim \boldsymbol{\pi}_\theta(\cdot \mid \boldsymbol{o}, z_b)$ on $\rho$
                \STATE $R_b \leftarrow \mathcal{R}(\tau_b)$
                \IF{$R_b > R^{*}$}
                    \STATE $\tau^{*} \leftarrow \tau_b$, \quad $R^{*} \leftarrow R_b$
                \ENDIF
            \ENDFOR
            \STATE Update $(\mu, \Sigma)$ using CMA-ES with $\{(z_b, R_b)\}_{b=1}^{B_{\max}}$
        \ENDWHILE
        \STATE \textbf{return} $\tau^{*}$
    \end{algorithmic}
\end{algorithm}

\section{Real-world examples where inference-time search is applicable}
\label{app:real-world}

Applying RL in real-world applications is a key driver of this research. Below, we describe three real-world cases where we have first-hand experience. In all of these, the user allows for a time budget, and a proxy of the solution’s quality is accessible to inform the search.

\paragraph{Printed Circuit Board Routing} The user submits the specification of a PCB board they want to route. This task can take up to several days or even weeks, even for an expert, depending on the complexity of the board. Therefore, users are typically satisfied with allowing an AI model to take at least a few minutes or even hours for search to obtain a better solution (routed board). Additionally, the user will often find it beneficial to leave more time and to grant more computation in order to improve the board further, which could then be easier to manufacture or reduce its production cost.

\paragraph{Bin Packing} When a logistics company wants to load containers with objects that need to be shipped overseas, operators typically have a few hours or days to plan which objects need to be transported, and can hence allow inference-time search. Here as well, the energy and money gains from having a better object dispatching (thus fewer containers to ship for the same amount of objects) will always justify a few hours of search, or the allowance of more computing spend.

\paragraph{Train Scheduling} Designing train schedules to meet a demand forecast, while proposing a feasible schedule (e.g., no collisions between trains), is another relevant example. This process usually happens once or twice a year and can take months to complete. Hence, there is no need to rely on a single rollout of the trained policy, and an inference-time search can instead be conducted over several days.

\section{Parameter counts for all base policy checkpoints}
\label{app:param-count}

Creating a COMPASS checkpoint from a pre-trained base policy requires adding new parameters to process the latent vectors (see~\cref{app:compass-details}), hence increasing the model size. However, this increase is small, never exceeding 2\% of the total base policy model size.

We report the parameter count for all base policy checkpoints and all corresponding COMPASS checkpoints for IPPO (\cref{tab:param-count-ippo}), MAPPO (\cref{tab:param-count-mappo}) and Sable (\cref{tab:param-count-sable}). We observe that (i) for IPPO and MAPPO, the increase in model size stays below 1.6\%, and (ii) for Sable, it typically adds 0.5\% to the total parameter count. Consequently, this model size increase has a minor impact on the memory and computational footprint of the method, which is confirmed by our study of the attempts achieved over time (reported in~\cref{app:inf-time-budget}).

\begin{table}[h!]
\centering
\caption{Parameter count comparison for MAPPO.}
\label{tab:param-count-mappo}
\begin{tabular}{lccc}
\toprule
\textbf{Scenario} & \textbf{Base Policy} & \textbf{COMPASS Policy} & \textbf{\% Increase} \\
\midrule
small-4ag-hard & 125061 & 127109 & 1.64\% \\
medium-6ag & 125317 & 127365 & 1.63\% \\
tiny-4ag-hard & 125061 & 127109 & 1.64\% \\
xlarge-4ag & 125061 & 127109 & 1.64\% \\
xlarge-4ag-hard & 125061 & 127109 & 1.64\% \\
large-8ag & 125573 & 127621 & 1.63\% \\
tiny-2ag-hard & 124805 & 126853 & 1.64\% \\
large-8ag-hard & 125573 & 127621 & 1.63\% \\
medium-4ag & 125061 & 127109 & 1.64\% \\
large-4ag & 125061 & 127109 & 1.64\% \\
large-4ag-hard & 125061 & 127109 & 1.64\% \\
medium-4ag-hard & 125061 & 127109 & 1.64\% \\
smacv2\_20\_units & 187417 & 189465 & 1.09\% \\
smacv2\_10\_units & 151567 & 153615 & 1.35\% \\
small-4ag & 125061 & 127109 & 1.64\% \\
con-10x10x10a & 124293 & 126341 & 1.65\% \\
con-15x15x23a & 125957 & 128005 & 1.63\% \\
\bottomrule
\end{tabular}
\end{table}

\begin{table}[h!]
\centering
\caption{Parameter count comparison for IPPO.}
\label{tab:param-count-ippo}
\begin{tabular}{lccc}
\toprule
\textbf{Scenario} & \textbf{Base Policy} & \textbf{COMPASS Policy} & \textbf{\% Increase} \\
\midrule
large-8ag-hard & 125573 & 127621 & 1.63\% \\
medium-6ag & 125317 & 127365 & 1.63\% \\
large-8ag & 125573 & 127621 & 1.63\% \\
tiny-4ag-hard & 125061 & 127109 & 1.64\% \\
small-4ag-hard & 125061 & 127109 & 1.64\% \\
tiny-2ag-hard & 124805 & 126853 & 1.64\% \\
xlarge-4ag-hard & 125061 & 127109 & 1.64\% \\
large-4ag-hard & 125061 & 127109 & 1.64\% \\
large-4ag & 125061 & 127109 & 1.64\% \\
xlarge-4ag & 125061 & 127109 & 1.64\% \\
medium-4ag & 125061 & 127109 & 1.64\% \\
medium-4ag-hard & 125061 & 127109 & 1.64\% \\
smacv2\_20\_units & 187417 & 189465 & 1.09\% \\
smacv2\_10\_units & 151567 & 153615 & 1.35\% \\
small-4ag & 125061 & 127109 & 1.64\% \\
con-10x10x10a & 124293 & 126341 & 1.65\% \\
con-15x15x23a & 125957 & 128005 & 1.63\% \\
\bottomrule
\end{tabular}
\end{table}

\begin{table}[h!]
\centering
\caption{Parameter count comparison for SABLE.}
\label{tab:param-count-sable}
\begin{tabular}{lccc}
\toprule
\textbf{Scenario} & \textbf{Base Policy} & \textbf{COMPASS Policy} & \textbf{\% Increase} \\
\midrule
large-8ag & 390096 & 392144 & 0.52\% \\
smacv2\_20\_units & 43363 & 43875 & 1.18\% \\
con-10x10x10a & 388422 & 390470 & 0.53\% \\
con-15x15x23a & 389907 & 391955 & 0.53\% \\
smacv2\_10\_units & 201435 & 202459 & 0.51\% \\
medium-6ag & 389454 & 391502 & 0.53\% \\
medium-4ag-hard & 389580 & 391628 & 0.53\% \\
large-4ag-hard & 389004 & 391052 & 0.53\% \\
large-8ag-hard & 389520 & 391568 & 0.53\% \\
small-4ag-hard & 389004 & 391052 & 0.53\% \\
tiny-4ag-hard & 1080524 & 1082572 & 0.19\% \\
small-4ag & 389196 & 391244 & 0.53\% \\
xlarge-4ag & 389004 & 391052 & 0.53\% \\
medium-4ag & 389580 & 391628 & 0.53\% \\
xlarge-4ag-hard & 1078796 & 1080844 & 0.19\% \\
large-4ag & 389004 & 391052 & 0.53\% \\
tiny-2ag-hard & 388746 & 390794 & 0.53\% \\
\bottomrule
\end{tabular}
\end{table}

\section{Discussion about the inference-time budget}
\label{app:inf-time-budget}

\paragraph{Choice of the budget used in the experiments} We create our experimental setting to be similar to practical use cases. Hence, we use time as the main budget constraint, instead of number of attempts (contrary to most literature on the topic). This enables to account for the cost of search and adaptation. Even though we always evaluate methods on \num{128} instances for each task, we ensure that each instance is solved fully independently, to avoid that the vectorisation process creates a bias towards one method or another. We choose to study methods with time budgets ranging from \num{30} to \num{300} seconds. The lower bound is taken to be small, to show that even short inference-time searches can provide significant benefit. The higher bound is chosen to be high enough to see which method can really benefit from a longer search budget. Concerning the compute capacity, we choose a range of values that seems achievable in practice. In real-world scenarios, simulations can get more complex and require one CPU core for one simulation. We hence believed that most use cases would fit within the limit of \num{256} simulations in parallel. Nevertheless, our code works for higher values and can be used to obtain results in new settings (both in terms of time and parallel attempts).

\paragraph{Total number of attempts achieved within time budget} To remain consistent with previous literature, and provide a better understanding of the search and adaptation cost of each inference strategy and each base policy, we provide the number of attempts achieved within the time limit, for all tasks and compute capacity (i.e., number of parallel attempts allowed). Values for \sable within 300 seconds are reported on~\cref{tab:budget_values} for \compass,  \cref{tab:fine_tuning_budget_values} for online fine-tuning, \cref{tab:budget_stochastic_values} for stochastic sampling. \sgbs builds new solutions from existing partial solutions, hence the count of the attempted solution cannot be obtained in the same way as other methods. 

Values for all base policies within 30 seconds are reported on~\cref{tab:budget_values_30sec},~\cref{tab:fine_tuning_budget_values_30} and~\cref{tab:budget_stochastic_values_transformed}.

\begin{table}[h]
\centering
\caption{Number of attempts (solutions generated) when using \sable + \compass during 300 seconds of inference-time search.}
\label{tab:budget_values}
\begin{tabular}{lrrrrrrr}
\toprule
\textbf{Task Name} 
& \multicolumn{7}{c}{\textbf{Number of attempts}} \\
\cmidrule(lr){2-8}
& \textbf{4} & \textbf{8} & \textbf{16} & \textbf{32} & \textbf{64} & \textbf{128} & \textbf{256} \\
\midrule
\texttt{con-10x10x10a} & 9960 & 19448 & 37520 & 67296 & 131456 & 248320 & 422144 \\
\texttt{con-15x15x23a} & 1608 & 3112 & 7744 & 14144 & 27456 & 53248 & 80896 \\
\texttt{large-4ag} & 2608 & 5208 & 9360 & 16896 & 30464 & 44032 & 59392 \\
\texttt{large-4ag-hard} & 2596 & 5000 & 9360 & 16832 & 29568 & 44288 & 59904 \\
\texttt{large-8ag} & 1612 & 3200 & 6240 & 10304 & 17280 & 23424 & 30976 \\
\texttt{large-8ag-hard} & 1708 & 3144 & 6096 & 10624 & 18880 & 29056 & 39680 \\
\texttt{medium-4ag} & 2088 & 4072 & 7904 & 15104 & 29376 & 48256 & 87296 \\
\texttt{medium-4ag-hard} & 2072 & 4040 & 7888 & 16992 & 29632 & 48384 & 87552 \\
\texttt{medium-6ag} & 1536 & 3016 & 5728 & 10816 & 21440 & 37760 & 74240 \\
\texttt{smacv2\_10\_units} & 7628 & 13248 & 26176 & 45088 & 74304 & 122624 & 174336 \\
\texttt{smacv2\_20\_units} & 5824 & 10428 & 18528 & 34112 & 64064 & 111232 & 180992 \\
\texttt{small-4ag} & 1964 & 3880 & 8672 & 14112 & 28096 & 61568 & 83456 \\
\texttt{small-4ag-hard} & 1876 & 4216 & 7104 & 15488 & 30464 & 52096 & 93696 \\
\texttt{tiny-2ag-hard} & 2736 & 6072 & 11856 & 22496 & 44544 & 77440 & 147200 \\
\texttt{tiny-4ag-hard} & 1384 & 2648 & 5024 & 10624 & 16640 & 32256 & 45568 \\
\texttt{xlarge-4ag} & 1816 & 3568 & 7952 & 15264 & 24320 & 45440 & 81152 \\
\texttt{xlarge-4ag-hard} & 1180 & 2264 & 5344 & 9696 & 17792 & 28544 & 51200 \\
\bottomrule
\end{tabular}
\end{table}

\begin{table}[h]
\centering
\caption{Number of attempts (solutions generated) when using online fine-tuning and \sable during 300 seconds of inference-time search.}
\label{tab:fine_tuning_budget_values}
\begin{tabular}{lrrrrrrr}
\toprule
\textbf{Task Name} & \multicolumn{7}{c}{\textbf{Number of attempts}} \\
\cmidrule(lr){2-8} 
& \textbf{4} & \textbf{8} & \textbf{16} & \textbf{32} & \textbf{64} & \textbf{128} & \textbf{256} \\
\midrule
\texttt{con-10x10x10a} & 10608 & 20256 & 37744 & 64448 & 115392 & 226560 & 395776 \\
\texttt{con-15x15x23a} & 2136 & 3952 & 6864 & 10496 & 16592 & 32000 & 60928 \\
\texttt{large-4ag} & 1792 & 3296 & 6560 & 9920 & 13376 & 24448 & 49664 \\
\texttt{large-4ag-hard} & 2064 & 3416 & 6448 & 8832 & 13248 & 24448 & 46592 \\
\texttt{large-8ag} & 1416 & 2336 & 4592 & 7552 & 11136 & 19200 & 34304 \\
\texttt{large-8ag-hard} & 1140 & 2576 & 4064 & 6752 & 12736 & 26624 & 44032 \\
\texttt{medium-4ag} & 2000 & 4176 & 7472 & 9920 & 15744 & 28160 & 48384 \\
\texttt{medium-4ag-hard} & 2024 & 3776 & 7232 & 10816 & 14848 & 28160 & 48384 \\
\texttt{medium-6ag} & 1700 & 2872 & 6448 & 8704 & 21120 & 32896 & 65792 \\
\texttt{smacv2\_10\_units} & 6848 & 12576 & 20176 & 37536 & 68160 & 156672 & 235776 \\
\texttt{smacv2\_20\_units} & 5700 & 10584 & 18176 & 25216 & 47488 & 104382 & 155904 \\
\texttt{small-4ag} & 1940 & 3688 & 7072 & 9600 & 14272 & 30336 & 49408 \\
\texttt{small-4ag-hard} & 1940 & 3688 & 7072 & 9600 & 14272 & 30336 & 49408 \\
\texttt{tiny-2ag-hard} & 2572 & 5600 & 9824 & 13344 & 22848 & 48768 & 86272 \\
\texttt{tiny-4ag-hard} & 2572 & 5568 & 9808 & 13344 & 22784 & 48768 & 86272 \\
\texttt{xlarge-4ag} & 1768 & 3856 & 6368 & 9792 & 13760 & 24448 & 46080 \\
\texttt{xlarge-4ag-hard} & 1280 & 2024 & 3776 & 6368 & 11456 & 21504 & 40192 \\
\bottomrule
\end{tabular}
\end{table}

\begin{table}[h]
\centering
\caption{Number of attempts (solutions generated) when sampling stochastically from \sable during 300 seconds of inference-time search.}
\label{tab:budget_stochastic_values}
\begin{tabular}{lrrrrrrr}
\toprule
\textbf{Task Name} 
& \multicolumn{7}{c}{\textbf{Number of attempts}} \\
\cmidrule(lr){2-8}
& \textbf{4} & \textbf{8} & \textbf{16} & \textbf{32} & \textbf{64} & \textbf{128} & \textbf{256} \\
\midrule
\texttt{con-10x10x10a} & 10632 & 20504 & 39524 & 70720 & 138624 & 264960 & 448512 \\
\texttt{con-15x15x23a} & 2048 & 4056 & 7840 & 14144 & 27648 & 54144 & 101376 \\
\texttt{large-4ag} & 2092 & 3624 & 7808 & 15456 & 30336 & 46464 & 82944 \\
\texttt{large-4ag-hard} & 2108 & 3696 & 8112 & 13344 & 26816 & 46592 & 82176 \\
\texttt{large-8ag} & 1508 & 2552 & 5824 & 9568 & 17984 & 30720 & 52480 \\
\texttt{large-8ag-hard} & 1324 & 2256 & 4432 & 8416 & 16576 & 34048 & 32208 \\
\texttt{medium-4ag} & 2388 & 4576 & 8064 & 15296 & 29760 & 53120 & 81664 \\
\texttt{medium-4ag-hard} & 2144 & 4616 & 7968 & 16992 & 29696 & 52992 & 88576 \\
\texttt{medium-6ag} & 1344 & 3496 & 6768 & 10848 & 25280 & 66048 & 66048 \\
\texttt{smacv2\_10\_units} & 7168 & 11056 & 25376 & 47088 & 94720 & 148608 & 261888 \\
\texttt{smacv2\_20\_units} & 4848 & 9144 & 17584 & 34688 & 55232 & 102144 & 228608 \\
\texttt{small-4ag} & 1472 & 3896 & 7072 & 16448 & 27840 & 61056 & 84224 \\
\texttt{small-4ag-hard} & 1812 & 3704 & 7072 & 16448 & 26880 & 59776 & 84224 \\
\texttt{tiny-2ag-hard} & 2824 & 6168 & 10432 & 20192 & 45504 & 77952 & 165632 \\
\texttt{tiny-4ag-hard} & 1600 & 2672 & 5888 & 9184 & 16960 & 33152 & 58880 \\
\texttt{xlarge-4ag} & 1832 & 3624 & 7888 & 13248 & 27584 & 50944 & 81152 \\
\texttt{xlarge-4ag-hard} & 1164 & 2280 & 4480 & 8064 & 15296 & 33152 & 58880 \\
\bottomrule
\end{tabular}
\end{table}

\begin{table}[h]
\centering
\caption{Number of attempts (solutions generated) when using \compass with \sable, \ippo and \mappo during 30 seconds of inference-time search.}
\label{tab:budget_values_30sec}
\begin{tabular}{lrrrrrrr|r}
\toprule
\textbf{Task Name} 
& \multicolumn{7}{c}{\textbf{Number of attempts - Sable}} 
& \multicolumn{1}{c}{\textbf{PPO}} \\
\cmidrule(lr){2-8} \cmidrule(lr){9-9}
& \textbf{4} & \textbf{8} & \textbf{16} & \textbf{32} & \textbf{64} & \textbf{128} & \textbf{256} & \textbf{64} \\
\midrule
\texttt{con-10x10x10a} & 996 & 1872  & 3744 & 6720  & 13120 & 24832 & 42240 & 65152 \\
\texttt{con-15x15x23a} & 160 & 384   & 768  & 1408  & 2752  & 5376  & 8192  & 36608 \\
\texttt{large-4ag}     & 260 & 464   & 928  & 1664  & 3072  & 4480  & 5888  & 5056  \\
\texttt{large-4ag-hard}& 256 & 464   & 928  & 1664 & 2944  & 4480  & 6144  &  4928  \\
\texttt{large-8ag}     & 160 & 312   & 624  & 1024  & 1728  & 2304  & 3072  & 3520  \\
\texttt{large-8ag-hard}& 168 & 304   & 608  & 1056  & 1856  & 2944  & 4096  & 3584  \\
\texttt{medium-4ag}    & 208 & 392   & 784  & 1504  & 2944  & 4864  & 8704  & 5120  \\
\texttt{medium-4ag-hard}&208 & 392   & 784  & 1696  & 2944  & 4864  & 8704  & 5056  \\
\texttt{medium-6ag}    & 152 & 288   & 576  & 1088  & 2112  & 3840  & 7424  & 4416  \\
\texttt{smacv2\_10\_units} & 760 & 1304 & 2608 & 4480 & 7424  & 12288 & 17408 & 30464 \\
\texttt{smacv2\_20\_units} & 584 & 928  & 1856 & 3392 & 6400  & 11136 & 18176 & 26880 \\
\texttt{small-4ag}     & 196 & 432   & 864  & 1408  & 2816  & 6144  & 8320  & 5248  \\
\texttt{small-4ag-hard}& 188 & 352   & 704  & 1536  & 3072  & 5120  & 9216  & 5248  \\
\texttt{tiny-2ag-hard} & 272 & 592   & 1184 & 2240  & 4480  & 7680  & 14592 & 6848  \\
\texttt{tiny-4ag-hard} & 136 & 248   & 496  & 1056  & 1664  & 3200  & 4608  & 5312  \\
\texttt{xlarge-4ag}    & 180 & 392   & 784  & 1536  & 2432  & 4480  & 8192  & 4160  \\
\texttt{xlarge-4ag-hard}&116 & 264   & 528  & 960   & 1792  & 2816  & 5120  & 4288  \\
\bottomrule
\end{tabular}
\end{table}

\begin{table}[h]
\centering
\caption{Number of attempts (solutions generated) when using stochastic sampling with \sable, \ippo and \mappo during 30 seconds of inference-time search.}
\label{tab:budget_stochastic_values_transformed}
\begin{tabular}{lrrrrrrr|r}
\toprule
\textbf{Task Name} 
& \multicolumn{7}{c}{\textbf{Number of attempts - Sable}} 
& \multicolumn{1}{c}{\textbf{PPO}} \\
\cmidrule(lr){2-8} \cmidrule(lr){9-9}
& \textbf{4} & \textbf{8} & \textbf{16} & \textbf{32} & \textbf{64} & \textbf{128} & \textbf{256} & \textbf{64} \\
\midrule
\texttt{con-10x10x10a} & 1060 & 2048 & 3940 & 7040 & 13760 & 26240 & 44800 & 103680 \\
\texttt{con-15x15x23a} & 204 & 400 & 780 & 1408 & 2752 & 5376 & 10112 & 51456 \\
\texttt{large-4ag} & 208 & 360 & 768 & 1536 & 3008 & 4608 & 8192 & 5312 \\
\texttt{large-4ag-hard} & 208 & 368 & 800 & 1312 & 2688 & 4608 & 8192 & 5184 \\
\texttt{large-8ag} & 148 & 248 & 576 & 928 & 1792 & 3072 & 5120 & 3904 \\
\texttt{large-8ag-hard} & 132 & 224 & 432 & 832 & 1664 & 3328 & 3200 & 3712 \\
\texttt{medium-4ag} & 236 & 456 & 800 & 1504 & 2944 & 5248 & 8128 & 5120 \\
\texttt{medium-4ag-hard} & 212 & 456 & 784 & 1696 & 2944 & 5248 & 8832 & 5312 \\
\texttt{medium-6ag} & 132 & 344 & 672 & 1056 & 2496 & 6528 & 6528 & 4480 \\
\texttt{smacv2\_10\_units} & 716 & 1104 & 2528 & 4704 & 9408 & 14848 & 26112 & 33408 \\
\texttt{smacv2\_20\_units} & 484 & 912 & 1744 & 3456 & 5504 & 10176 & 22784 & 31808 \\
\texttt{small-4ag} & 144 & 384 & 704 & 1632 & 2752 & 6016 & 8384 & 5184 \\
\texttt{small-4ag-hard} & 180 & 368 & 704 & 1632 & 2688 & 5888 & 8384 & 5248 \\
\texttt{tiny-2ag-hard} & 280 & 608 & 1040 & 2016 & 4544 & 7808 & 16512 & 4544 \\
\texttt{tiny-4ag-hard} & 160 & 264 & 576 & 896 & 1664 & 3328 & 5888 & 5312 \\
\texttt{xlarge-4ag} & 180 & 360 & 784 & 1312 & 2752 & 5120 & 8128 & 4416 \\
\texttt{xlarge-4ag-hard} & 116 & 224 & 448 & 800 & 1536 & 3328 & 5888 & 4352 \\
\bottomrule
\end{tabular}
\end{table}

\begin{table}[h]
\centering
\caption{Number of attempts (solutions generated) when using online fine-tuning with \sable, \ippo and \mappo during 30 seconds of inference-time search.}
\label{tab:fine_tuning_budget_values_30}
\begin{tabular}{lrrrrrrr|r}
\toprule
\textbf{Task Name} 
& \multicolumn{7}{c}{\textbf{Number of attempts - Sable}} 
& \multicolumn{1}{c}{\textbf{PPO}} \\
\cmidrule(lr){2-8} \cmidrule(lr){9-9}
& \textbf{4} & \textbf{8} & \textbf{16} & \textbf{32} & \textbf{64} & \textbf{128} & \textbf{256} & \textbf{64} \\
\midrule
\texttt{con-10x10x10a}     & 1060 & 2024 & 3760 & 6432 & 11520 & 22656 & 39424 & 23680 \\
\texttt{con-15x15x23a}     & 212  & 392  & 672  & 1024 & 1600  & 3200  & 5888  & 5888 \\
\texttt{large-4ag}         & 176  & 328  & 656  & 992  & 1280  & 2432  & 4864  & 3136 \\
\texttt{large-4ag-hard}    & 204  & 336  & 640  & 864  & 1280  & 2432  & 4608  & 2944 \\
\texttt{large-8ag}         & 140  & 232  & 448  & 736  & 1088  & 1920  & 3328  & 2304 \\
\texttt{large-8ag-hard}    & 112  & 256  & 400  & 672  & 1216  & 2560  & 4352  & 2112 \\
\texttt{medium-4ag}        & 200  & 416  & 736  & 992  & 1536  & 2816  & 4608  & 2880 \\
\texttt{medium-4ag-hard}   & 200  & 376  & 720  & 1056 & 1472  & 2816  & 4608  & 2880 \\
\texttt{medium-6ag}        & 168  & 280  & 640  & 864  & 2112  & 3200  & 6400  & 4160 \\
\texttt{smacv2\_10\_units} & 684  & 1256 & 2016 & 3744 & 6784  & 15616 & 23552 & 8960 \\
\texttt{smacv2\_20\_units} & 568  & 1056 & 1808 & 2496 & 4736  & 10368 & 15360 & 7936 \\
\texttt{small-4ag}         & 192  & 368  & 704  & 960  & 1408  & 2944  & 4864  & 2816 \\
\texttt{small-4ag-hard}    & 192  & 368  & 704  & 960  & 1408  & 2944  & 4864  & 2816 \\
\texttt{tiny-2ag-hard}     & 256  & 560  & 976  & 1312 & 2240  & 4864  & 8448  & 3904 \\
\texttt{tiny-4ag-hard}     & 256  & 552  & 976  & 1312 & 2240  & 4864  & 8448  & 3904 \\
\texttt{xlarge-4ag}        & 176  & 384  & 624  & 960  & 1344  & 2432  & 4608  & 2944 \\
\texttt{xlarge-4ag-hard}   & 128  & 200  & 368  & 608  & 1088  & 2048  & 3840  & 2432 \\
\bottomrule
\end{tabular}
\end{table}

\section{Additional discussion about COMPASS}
\label{app:why-compass-better}

In~\cref{sec:scaling-budget} we mention explanations to why \compass gets good performance and scaling trends. In this section, we elaborate on these, and highlight three main aspects.

First, \compass trains a “multiple attempts” objective, unlike other methods, which are training for “single attempt” performance. Thus, \compass’s training objective anticipates that what matters is the maximum performance over multiple attempts, rather than the average performance over these attempts. While its average solution quality may be lower, it has a higher probability of producing one excellent solution. This makes it inherently better suited for inference-time search where multiple attempts are allowed.

Second, \compass creates a dense space of diverse and specialised policies, hence creating a strong exploration capacity. While most other methods rely on the entropy of the action distribution to explore (i.e. sampling stochastically from the pre-trained policy), \compass can use the wide diversity encoded in its latent space to build much more diverse policies.

Finally, the Covariance Matrix Adaptation mechanism (\cmaes) used by \compass at inference time enables it to navigate the latent space and identify the most adapted (suitable) latent for the problem instance at hand. Using a \cmaes search in a 16-dimensional space is more efficient and less prone to getting stuck in local optima compared to using gradient descent in the entire policy parameter space (i.e. online fine-tuning).

\section{Additional experimental details and error bars}
\label{app:additional-error-bars}

\begin{figure}[h!]
    \centering
    \vspace{-1.5\baselineskip}
    \includegraphics[width=0.95\textwidth]{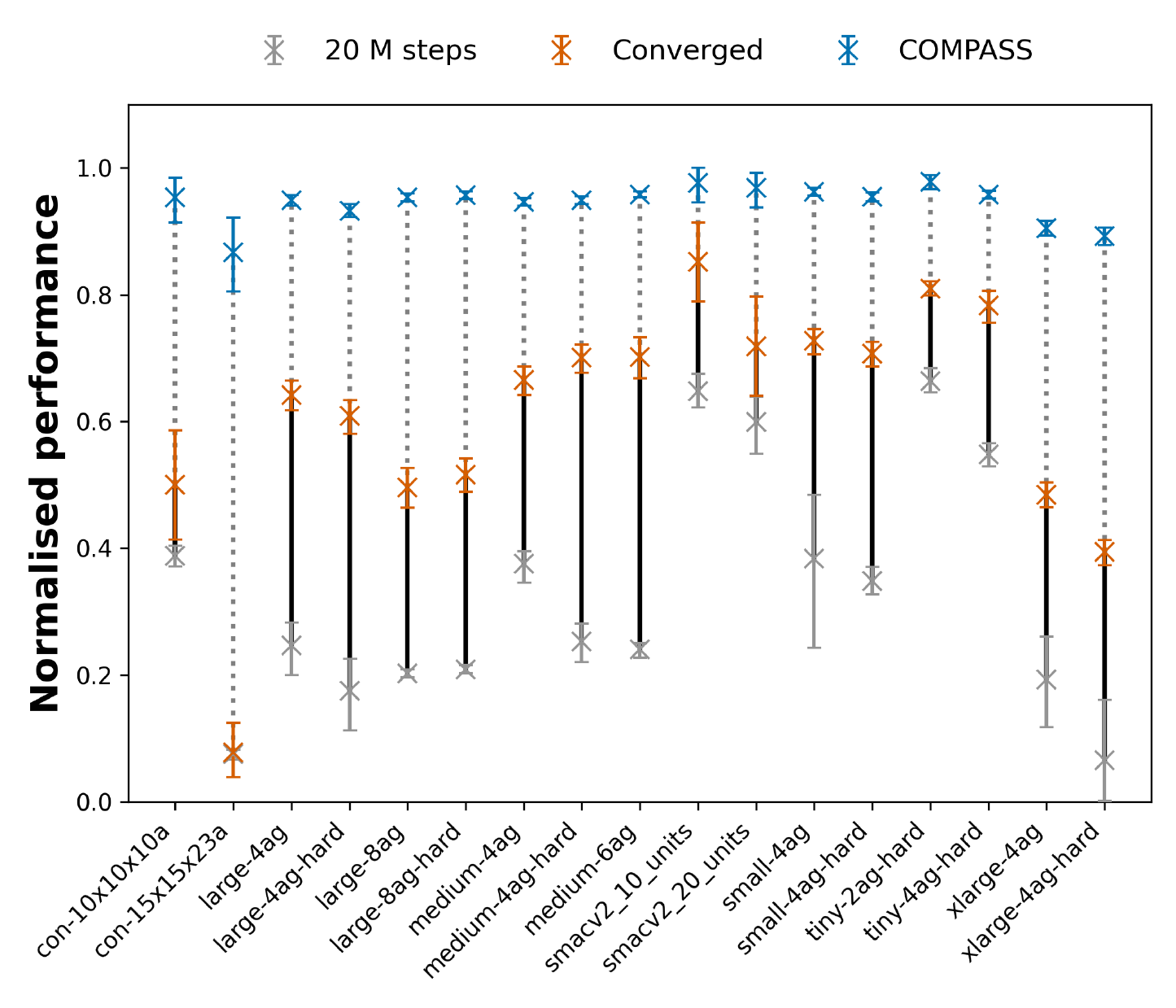}
    \caption{\textbf{Training the SOTA algorithm Sable to convergence} is not enough to achieve optimal zero-shot performance, but using 30 seconds of search enables to consistenly reach above 85\% normalised performance across the whole benchmark. We report the mean and 95\% bootstrap confidence intervals.}
    \label{fig:zero-shot-cvg-errors}
    \vspace{-0.5\baselineskip}
\end{figure}

In~\cref{fig:zero-shot-cvg}, we report the zero-shot performance of SOTA method \sable, and compare it with the best results observed for 30 seconds of inference-time search. We report the error values in~\cref{tab:error-bars-fig4} and the figure with error bars in~\cref{fig:zero-shot-cvg-errors}. 

We report the mean and 95\% bootstrap confidence intervals. The 20M timestep values are aggregated over 10 independent runs and computed directly from the benchmarking data provided along with the Sable paper~\citep{mahjoub2025sable}, for the policy trained to convergence and for COMPASS, we give the results aggregated over 128 independent runs. 

\begin{table}[h!]
\centering
\caption{Performance comparison between the base policy (after 20M timesteps and after convergence) and COMPASS. Values indicate mean performance with 95\% confidence intervals.}
\label{tab:error-bars-fig4}
\begin{tabular}{lccc}
\toprule
\textbf{Scenario} & \textbf{20M Timesteps} & \textbf{Converged} & \textbf{COMPASS} \\
\midrule
con-10x10x10a & 0.388 (0.372, 0.404) & 0.500 (0.414, 0.586) & 0.953 (0.914, 0.984) \\
con-15x15x23a & 0.075 (0.067, 0.083) & 0.078 (0.039, 0.125) & 0.867 (0.805, 0.922) \\
large-4ag & 0.246 (0.200, 0.283) & 0.642 (0.618, 0.664) & 0.949 (0.940, 0.957) \\
large-4ag-hard & 0.175 (0.113, 0.225) & 0.609 (0.581, 0.633) & 0.933 (0.922, 0.943) \\
large-8ag & 0.203 (0.197, 0.208) & 0.496 (0.464, 0.526) & 0.953 (0.947, 0.959) \\
large-8ag-hard & 0.209 (0.202, 0.216) & 0.516 (0.489, 0.541) & 0.957 (0.950, 0.963) \\
medium-4ag & 0.376 (0.346, 0.396) & 0.666 (0.642, 0.686) & 0.947 (0.941, 0.953) \\
medium-4ag-hard & 0.253 (0.220, 0.282) & 0.701 (0.677, 0.721) & 0.949 (0.943, 0.955) \\
medium-6ag & 0.240 (0.227, 0.250) & 0.702 (0.668, 0.733) & 0.958 (0.954, 0.963) \\
smacv2\_10\_units & 0.648 (0.623, 0.675) & 0.852 (0.789, 0.914) & 0.977 (0.945, 1.000) \\
smacv2\_20\_units & 0.599 (0.548, 0.639) & 0.719 (0.641, 0.797) & 0.969 (0.938, 0.992) \\
small-4ag & 0.384 (0.243, 0.484) & 0.728 (0.706, 0.746) & 0.963 (0.956, 0.969) \\
small-4ag-hard & 0.349 (0.327, 0.371) & 0.707 (0.687, 0.725) & 0.955 (0.948, 0.962) \\
tiny-2ag-hard & 0.664 (0.646, 0.684) & 0.810 (0.799, 0.821) & 0.978 (0.967, 0.989) \\
tiny-4ag-hard & 0.548 (0.529, 0.566) & 0.783 (0.756, 0.806) & 0.958 (0.951, 0.964) \\
xlarge-4ag & 0.193 (0.118, 0.261) & 0.485 (0.465, 0.504) & 0.905 (0.893, 0.916) \\
xlarge-4ag-hard & 0.065 (0.001, 0.161) & 0.394 (0.373, 0.413) & 0.892 (0.879, 0.906) \\
\bottomrule
\end{tabular}
\end{table}

\end{document}